\documentclass{article}

\usepackage{arxiv}

\usepackage[utf8]{inputenc} 
\usepackage[T1]{fontenc}    
\usepackage[backref=page]{hyperref}       
\usepackage{url}            
\usepackage{booktabs}       
\usepackage{amsfonts}       
\usepackage{nicefrac}       
\usepackage{microtype}      
\usepackage{lipsum}         
\usepackage{graphicx}
\usepackage{natbib}
\usepackage{doi}
\usepackage{amsmath}        
\usepackage[nameinlink,capitalise,noabbrev]{cleveref}       
\usepackage{chngcntr}       
\usepackage{pdflscape}      
\usepackage{rotating}       
\usepackage{makecell}       
\usepackage[labelfont=bf]{caption} 
\usepackage{float}          

\title{Fully transformer-based biomarker prediction from colorectal cancer histology: a large-scale multicentric study}

\date{%
\protect\tiny
\textsuperscript{1}Helmholtz Munich – German Research Center for Environment and Health, Munich, Germany.
\textsuperscript{2}School of Computation, Information and Technology, Technical University of Munich, Munich, Germany.
\textsuperscript{3}Else Kroener Fresenius Center for Digital Health (EFFZ), Technical University Dresden, Dresden, Germany.
\textsuperscript{4}Division of Pathology and Data Analytics, Leeds Institute of Medical Research at St James's, University of Leeds, Leeds, United Kingdom.
\textsuperscript{5}Department of Pathology, GROW School for Oncology and Developmental Biology, Maastricht University Medical Center+, Maastricht, The Netherlands.
\textsuperscript{6}Department of Epidemiology, Maastricht University Medical Center+, Maastricht, The Netherlands.
\textsuperscript{7}Division of Clinical Epidemiology and Aging Research, German Cancer Research Center (DKFZ), Heidelberg, Germany.
\textsuperscript{8}Institute of Pathology und Molecular Pathology, Johannes Kepler University Hospital Linz, Linz, Österreich.
\textsuperscript{9}Gemeinschaftspraxis Pathologie, Starnberg, Germany.
\textsuperscript{{10}}Nuffield Department of Population Health, University of Oxford, Oxford, United Kingdom.
\textsuperscript{{11}}Center for Precision Medicine and Department of Medical Oncology, City of Hope National Medical Center, Duarte, CA, USA.
\textsuperscript{{12}}Department of Pathology, City of Hope Comprehensive Cancer Center, Duarte, CA, USA.
\textsuperscript{{13}}Department of Community Medicine \& Epidemiology, Lady Davis Carmel Medical Center, Ruth \& Bruce Rappaport Faculty of Medicine, Technion-Israel Institute of Technology, Israel.
\textsuperscript{{14}}Steve and Cindy Rasmussen Institute for Genomic Medicine, Lady Davis Carmel Medical Center and Technion Faculty of Medicine, Clalit National Cancer Control Center, Haifa, Israel.
\textsuperscript{{15}}Precision Medicine Centre of Excellence, Health Sciences Building, The Patrick G Johnston Centre for Cancer Research, Queen's University Belfast, Belfast, United Kingdom.
\textsuperscript{{16}}Regional Molecular Diagnostic Service, Belfast Health and Social Care Trust, Belfast, United Kingdom.
\textsuperscript{{17}}The Patrick G Johnston Centre for Cancer Research, Queen's University Belfast, Belfast, United Kingdom.
\textsuperscript{{18}}Department of Cellular Pathology, Belfast Health and Social Care Trust, Belfast, United Kingdom.
\textsuperscript{{19}}Centre for Public Health, Queen's University Belfast, Belfast, United Kingdom.
\textsuperscript{{20}}Integrated Pathology Unit, Institute for Cancer Research and Royal Marsden Hospital, London, United Kingdom.
\textsuperscript{{21}}Division of Preventive Oncology, German Cancer Research Center (DKFZ) and National Center for Tumor Diseases (NCT), Heidelberg, Germany.
\textsuperscript{{22}}German Cancer Consortium (DKTK), German Cancer Research Center (DKFZ), Heidelberg, Germany.
\textsuperscript{{23}}Department of Diagnostic and Interventional Radiology, University Hospital RWTH Aachen, Aachen, Germany.
\textsuperscript{{24}}School of Biomedical Engineering and Imaging Sciences, King’s College London, London, United Kingdom.
\textsuperscript{{25}}Institute of Pathology, Technical University Munich, Munich, Germany.
\textsuperscript{{26}}Institute of Pathology Munich-North, Munich, Germany.
\textsuperscript{{27}}Medical Oncology, National Center for Tumor Diseases (NCT), University Hospital Heidelberg, Heidelberg. \\
}

\author{Sophia J. Wagner\textsuperscript{1,2,3}, \hspace{-10pt}
\And Daniel Reisenbüchler\textsuperscript{1}, \hspace{-10pt}
\And Nicholas P. West\textsuperscript{4}, \hspace{-10pt}
\And Jan Moritz Niehues\textsuperscript{3}, \hspace{-10pt}
\And Gregory Patrick Veldhuizen\textsuperscript{3}, \hspace{-10pt}
\And Philip Quirke\textsuperscript{4}, \hspace{-10pt}
\And Heike I. Grabsch\textsuperscript{4,5}, \hspace{-10pt}
\And Piet A. van den Brandt\textsuperscript{6}, \hspace{-10pt}
\And Gordon G. A. Hutchins\textsuperscript{4}, \hspace{-10pt}
\And Susan D. Richman\textsuperscript{4}, \hspace{-10pt}
\And Tanwei Yuan\textsuperscript{7}, \hspace{-10pt}
\And Rupert Langer\textsuperscript{8}, \hspace{-10pt}
\And Josien Christina Anna Jenniskens\textsuperscript{6}, \hspace{-10pt}
\And Kelly Offermans\textsuperscript{6}, \hspace{-10pt}
\And Wolfram Mueller\textsuperscript{9}, \hspace{-10pt}
\And Richard Gray\textsuperscript{10}, \hspace{-10pt}
\And Stephen B. Gruber\textsuperscript{11}, \hspace{-10pt}
\And Joel K. Greenson\textsuperscript{12}, \hspace{-10pt}
\And Gad Rennert\textsuperscript{13,14}, \hspace{-10pt}
\And Joseph D. Bonner\textsuperscript{13}, \hspace{-10pt}
\And Daniel Schmolze\textsuperscript{11}, \hspace{-10pt}
\And Jacqueline A. James\textsuperscript{15,16,17}, \hspace{-10pt}
\And Maurice B. Loughrey\textsuperscript{17,18,19}, \hspace{-10pt}
\And Manuel Salto-Tellez\textsuperscript{15,16,20}, \hspace{-10pt}
\And Hermann Brenner\textsuperscript{7,21,22}, \hspace{-10pt}
\And Michael Hoffmeister\textsuperscript{7}, \hspace{-10pt}
\And Daniel Truhn\textsuperscript{23}, \hspace{-10pt}
\And Julia A. Schnabel\textsuperscript{1,2,24}, \hspace{-10pt}
\And Melanie Boxberg\textsuperscript{25,26}, \hspace{-10pt} 
\And Tingying Peng\thanks{Last authors contributed equally}\textsuperscript{ \ ,1}, \hspace{-10pt}
\And Jakob Nikolas Kather\textsuperscript{$*$,3,4,2}
}

\renewcommand{\shorttitle}{Fully transformer-based biomarker prediction from colorectal cancer histology: a large-scale multicentric study}

\hypersetup{
pdftitle={A template for the arxiv style},
pdfsubject={q-bio.NC, q-bio.QM},
pdfauthor={David S.~Hippocampus, Elias D.~Striatum},
pdfkeywords={First keyword, Second keyword, More},
}

\begin{document}
\maketitle

\begin{abstract}
	\textbf{Background:} Deep learning (DL) can extract predictive and prognostic biomarkers from routine pathology slides in colorectal cancer. For example, a DL test for the diagnosis of microsatellite instability (MSI) in CRC has been approved in 2022. Current approaches rely on convolutional neural networks (CNNs). Transformer networks are outperforming CNNs and are replacing them in many applications, but have not been used for biomarker prediction in cancer at a large scale. In addition, most DL approaches have been trained on small patient cohorts, which limits their clinical utility.

    \textbf{Methods:} In this study, we developed a new fully transformer-based pipeline for end-to-end biomarker prediction from pathology slides. We combine a pre-trained transformer encoder and a transformer network for patch aggregation, capable of yielding single and multi-target prediction at patient level. We train our pipeline on over 9,000 patients from 10 colorectal cancer cohorts.

    \textbf{Results:} A fully transformer-based approach massively improves the performance, generalizability, data efficiency, and interpretability as compared with current state-of-the-art algorithms. After training on a large multicenter cohort, we achieve a sensitivity of 0.97 with a negative predictive value of 0.99 for MSI prediction on surgical resection specimens. We demonstrate for the first time that resection specimen-only training reaches clinical-grade performance on endoscopic biopsy tissue, solving a long-standing diagnostic problem. 

    \textbf{Interpretation:} A fully transformer-based end-to-end pipeline trained on thousands of pathology slides yields clinical-grade performance for biomarker prediction on surgical resections and biopsies. Our new methods are freely available under an open source license.
\end{abstract}


\section{Introduction}

Precision oncology in colorectal cancer (CRC) requires the evaluation of genetic biomarkers, such as microsatellite instability (MSI)  \citep{Kather2019-zm,Cao2020-uv,Echle2020-ot,Bilal2021-st,Lee2021-fz,Schirris2022-co,Schrammen2022-ly,Yamashita2021-on} and mutations in the \textit{BRAF} \citep{Bilal2021-st,Schrammen2022-ly} \textit{NRAS}/\textit{KRAS} \citep{Jang2020-ua} genes. These biomarkers are typically measured by polymerase chain reaction (PCR), sequencing (for \textit{BRAF} and \textit{KRAS}), or immunohistochemical assays. Biomarker identification in patients with CRC is an important step in providing treatment as recommended by various medical guidelines, such as those in the USA (NCCN guideline) \citep{Benson2018-za}, UK (NICE guideline) \citep{NICEguideline}, and EU (ESMO guideline) \citep{Schmoll2012-mg}. Increasingly, genetic biomarkers such as MSI are also used in earlier tumor stages of CRC \citep{Chalabi2022-tp}. In the future, the importance of biomarker-stratified therapy will likely increase \citep{Vacante2018-yn}. The presence of MSI should also trigger additional diagnostic processes for Lynch syndrome, one of the most prevalent hereditary cancer syndromes. However, genetic diagnostic assays have several disadvantages. For many patients in low- and middle-income countries, genetic biomarkers are not routinely available due to the prohibitive costs and complex infrastructure required for testing. Even in high-income countries with universal healthcare coverage where genetic biomarkers may be routinely available, their utilization is not without its drawbacks. In such contexts, biomarker assessment can take several days to weeks delaying therapy decisions \citep{Lim2015-yt}.

The diagnosis of CRC requires a pathologist's histopathological evaluation of tissue sections. Thus, tissue sections stained with hematoxylin and eosin (H\&E) are routinely available for all patients with CRC. Since 2019, dozens of studies have shown that Deep Learning (DL) can predict genetic biomarkers directly from digitized H\&E-stained CRC tissue sections  \citep{Kather2019-zm,Echle2020-ot}. Based on these studies, the first commercial DL algorithm for biomarker detection from H\&E images has been approved for routine clinical use in Europe in 2022 (MSIntuit, Owkin, Paris/New York).  When evaluated in external patient cohorts, the state-of-the-art approaches have reached a sensitivity and specificity of 0.95 and 0.46, respectively \citep{Svrcek2022-ck}. The rather low specificity is a key limitation of established approaches. Another clinically significant limitation of current approaches is the poor performance on endoscopic biopsy tissue. Recent clinical trials (NICHE \citep{Chalabi2020-pf} and NICHE-2 \citep{Chalabi2022-tp}) showed high efficacy of neoadjuvant immunotherapy for patients with MSI CRC. These findings imply that in the future every CRC patient should be tested for MSI on the initial biopsy tissue, although not all current medical guidelines reflect this. Only \citet{Echle2020-ot} determined the performance of DL-based biomarker prediction on CRC biopsy tissue and reported an even lower performance on biopsy tissue than on surgical resection tissue sections (biopsy AUROC: 0.78; resection AUROC of 0.96). Current clinically approved commercial products for MSI detection in CRC from histopathology are only applicable to surgical resection tissue. 

The technology underlying these algorithms in literature is based on weakly-supervised learning, consisting of two components: the feature extractor and the aggregation module \citep{Bilal2022-rt}. The feature extractor is mostly based on a convolutional neural network (CNN), which processes multiple small tissue regions called tiles or patches. The CNN-based representations obtained from these tiles are subsequently aggregated to obtain a single prediction for the patient. Between 2019 and 2021, most studies used simple heuristics, such as taking the maximum (max pooling) or averaging (average pooling), as an aggregation module. Since then, variations of multiple instance learning (MIL) have become the new standard for this task \citep{Ilse2018-it,Schirris2022-co}. The most common approach replaces the pooling aggregation with a small two-layer network to learn the patch-level weighting of the embeddings \citep{Ilse2018-it}. However, current MIL approaches univariately consider a single tile during aggregation and do not place it in context with other tiles even though local and global contexts are crucial for medical diagnosis.

In many non-medical and medical image processing tasks, transformer neural networks have recently been adopted for computer vision tasks \citep{Dosovitskiy2020-gz,Liu2021-up,He2022-mg}, replacing CNNs because of their improved performance and robustness \citep{Ghaffari_Laleh2022-sr}. Originally proposed for sequencing tasks such as natural language processing, transformer networks show impressive capabilities of learning long-range dependencies and contextualizing concepts in long sequences. In computational pathology, transformers have therefore been proposed as potentially superior feature extractors \citep{Wang2022-fd} or aggregation models \citep{Chen2022-ej,Shao2021-vs,Reisenbuchler2022-dk,Ghaffari_Laleh2022-ic}, though these proposals still lack empirical evidence from large-scale analyses. 

In this work, we first aim to enhance the performance of DL-based biomarker detection from pathology slides. Thereafter, in order to provide large-scale evidence of the performance on clinically relevant tasks, we investigate the use of a fully transformer-based workflow in CRC. Here, we present a new method derived from a transformer-based feature extractor and a transformer-based aggregation model, which we evaluate in a large multi-centric study of 10 cohorts with resection specimen slides from over 9,000 CRC patients worldwide, as well as one large cohort of CRC biopsies from over 1,500 patients. 

\section{Methods}

\begin{figure}[ht]
    \centering
    \includegraphics[width=\textwidth]{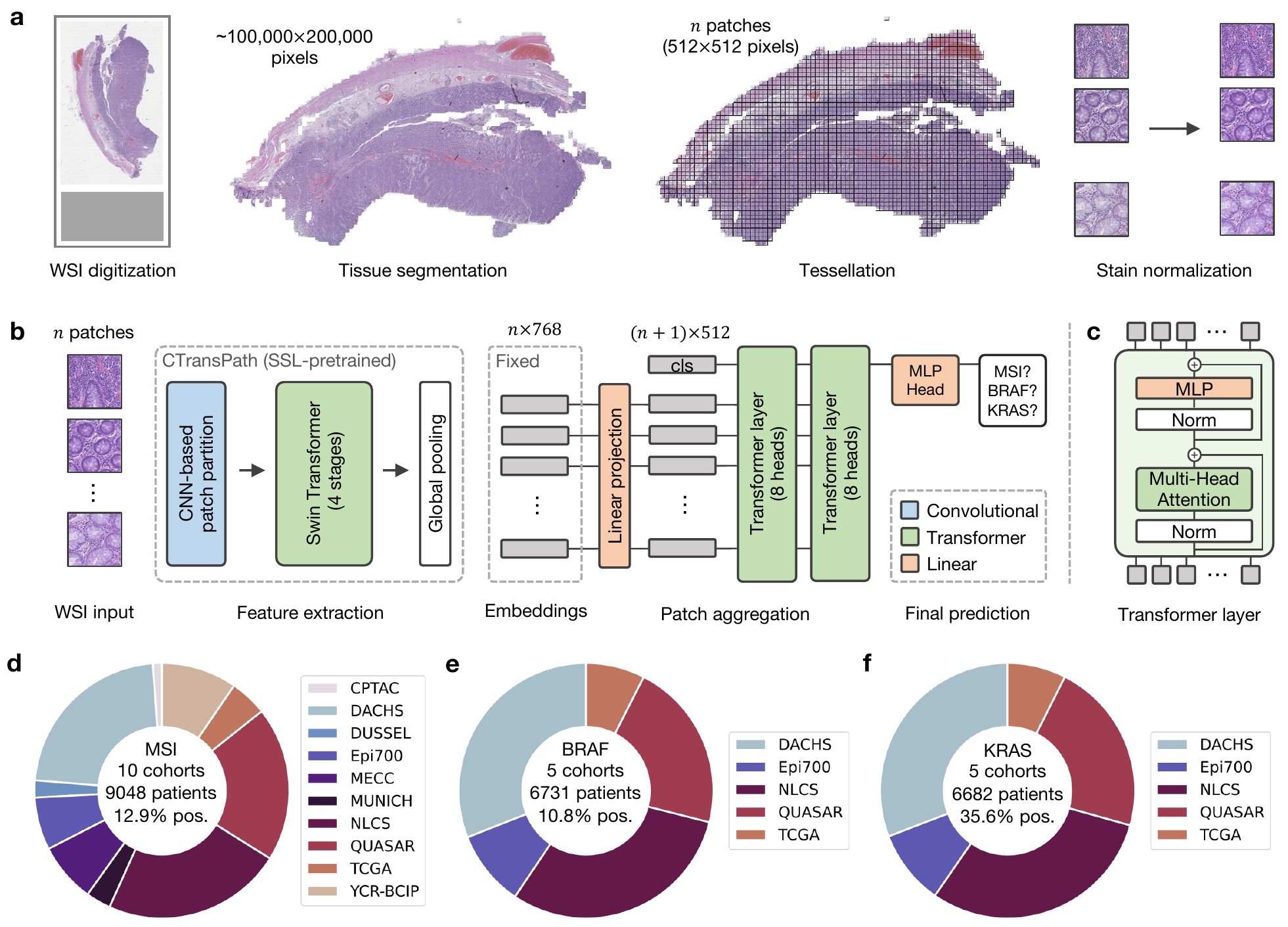}
    \caption{\textbf{Workflow with pre-processing, model architecture and cohort overview.} a) The data pre-processing pipeline with the steps whole slide image (WSI) digitization, tissue segmentation, tessellation into tiles, and stain normalization, b) the model architecture including the pre-trained feature extractor CTransPath and our transformer-based aggregation modules, and c) details of the transformer layer architecture. d) Overview of the 10 MSI cohorts used in this study with e) and f) the subsets of five cohorts with \textit{BRAF} and \textit{KRAS} annotations, respectively.}
    \label{fig:model}
\end{figure}

\subsection{Model description}
Our biomarker prediction pipeline consists of three steps (\cref{fig:model}): i) the data pre-processing pipeline (\cref{fig:model}a), ii) the transformer-based feature extractor, and iii) the transformer-based aggregation module that yields the final prediction from the embeddings of all patches of a whole-slide image (WSI) (\cref{fig:model}b).

In the pre-processing pipeline, tissue regions are segmented using RGB thresholding and Canny edge detection \citep{Canny1986-uw} to detect background and blurry regions. We include all tiles from a WSI, i.e., both tumor and healthy tissue tiles, thus reducing the burden of manual annotations when applying the algorithm. Subsequently, the WSI is tessellated into tiles of size 512$\times$512 pixels at 20$\times$ magnification with a resolution of 0.5 microns per pixel. To reduce the impact of the staining color on the model generalization, all tiles are stain-normalized with Macenko’s color deconvolution algorithm \citep{Macenko2009-an} using a tile from the DACHS cohort as a reference image (\cref{fig:normalization_template}). 

We extract feature representations of dimension 768 for every tile using the CTransPath model \citep{Wang2022-fd}. (\cref{fig:model}b). The model architecture is based on a Swin Transformer \citep{Liu2021-up} that combines the hierarchical structure of CNNs with the global self-attention modules of transformers by computing self-attention in a sliding-window fashion. Similar to CNNs, these are stacked to increase the receptive field in every stage. CTransPath consists of three convolutional layers at the beginning to facilitate local feature extraction and improve training stability \citep{Wang2022-fd}, followed by four Swin Transformer stages. \citet{Wang2022-fd} trained the network using an unsupervised contrastive loss on data from TCGA and PAIP \citep{Kim2021-bw} from multiple organs and provided the weights for public use. The embeddings for each tile are stored for the subsequent training procedure.

The final part of the model takes all patches of a WSI as input and predicts one biomarker for all input patches in a weakly supervised manner (\cref{fig:model}b). Common attention-based MIL approaches \citep{Ilse2018-it} use a small neural network, which mostly consists of two layers, to compute patch importance based on the embeddings. Each weight is computed based on one patch and finally, all weights are normalized over the input elements. In contrast to this, in our model, the patch embeddings are passed into a transformer network using multi-headed self-attention that considers the patch embeddings as a sequence and relates each element to every other element. In particular, assuming that $x\in \mathbb{R}^{n\times d}$ is the input sequence representing a WSI with $n$ patch embeddings of dimension $d$, the self-attention (SA) layer computes a query-key product in the following way
\begin{align}
    \mathrm{SA}(Q,K,V) = \mathrm{softmax}\left(\frac{QK^T}{\sqrt{d_k}}\right)V,
\end{align}
where queries $Q\in \mathbb{R}^{n\times d_k}$, keys $K\in \mathbb{R}^{n\times d_k}$, and values $V\in \mathbb{R}^{n\times d_v}$ .These are computed from the input sequence $x$ by 
\begin{align}
     Q = W_Q \cdot x, \quad K = W_K \cdot x, \quad V = W_V \cdot x,
\end{align}
where  $W_Q\in \mathbb{R}^{d\times d_k}$, $W_K\in \mathbb{R}^{d\times d_k}$, and $W_V\in \mathbb{R}^{d\times d_v}$ are learnable parameters. Multi-headed self-attention (MSA) applies self-attention in every head and concatenates the heads in a weighted manner:
\begin{align}
    \mathrm{MSA}(x) & = \mathrm{concat}(\mathrm{head}_1, ..., \mathrm{head}_h)\cdot W_O, \\
    \text{where} \ \mathrm{head}_i & =\mathrm{SA}(Q^{(i)},K^{(i)},V^{(i)}) \ \text{for} \ i \in \{1, ..., h\}
\end{align}
and $W_O\in \mathbb{R}^{hd_v\times d}$ is learnable. We choose a small transformer network architecture consisting of two layers with each eight heads ($h=8$), a latent dimension of $512$, and the same dimension for query, keys, and values. Therefore, the latent dimension of each head is $d_v=d_k=64$, such that $hd_v=8\cdot 64=512$.

Assuming that $n$ is the number of patches per WSI, the embeddings of each patch $i \in \{1, \dots, n\}$ are stacked to a sequence of dimension $n \times 768$ and are passed through a linear projection layer followed by the non-linear activation ReLU to reduce the dimension from $768$ to $512$. Subsequently, a class token is concatenated to the input, similar to the usage in vision transformers \citep{Dosovitskiy2020-gz}, yielding an input of dimension $(n+1) \times 512$ that is passed to the transformer layer. In each transformer layer, a block of layer normalization and multi-headed self-attention is followed by a block of layer normalization and a multi-layer perceptron (MLP), with skip connections applied across each block (\cref{fig:model}c). 

After the two transformer layers, the class token of size $1 \times 512$ is passed into an MLP head. Depending on the number of class tokens used, this enables single-target or multi-target binary prediction. Instead of attaching a class token, all $n$ sequence elements could be averaged to a single sequence element of size $1 \times 512$ and passed into the MLP head. The averaging approach achieves similar performance to the class token version (\cref{tab:results}), but we decided to use the class token for better interpretability of the attention heads. We also compared our model architecture to the existing transformer-based aggregation module TransMIL \citep{Shao2021-vs} (\cref{tab:results}).

\subsection{Ethics statement}
In this study, we retrospectively analyzed anonymized patient samples from multiple academic institutions. At each of the following sites, the respective ethics board has given consent to this analysis: DACHS, Epi700, MECC, MUNICH, NLCS, and QUASAR. At the following sites, specific ethics approval was not required for a retrospective analysis of anonymized samples: CPTAC, DUSSEL, TCGA, and YCR-BCIP. Our study adheres to STARD (\cref{tab:stard}).

\subsection{Cohort description}
Through coordination by the MSIDETECT consortium (\href{www.msidetect.eu}{www.msidetect.eu}), we have collected H\&E tissue sections of 9,048 patients with CRC from 10 patient cohorts in total, including two public databases (\cref{fig:model}d-f). The cohorts obtained are as follows:
\begin{enumerate}
    \item The public database “The Clinical Proteomic Tumor Analysis Consortium”, CPTAC (publicly available at \href{https://pdc.cancer.gov/pdc/}{https://pdc.cancer.gov/pdc/}, USA) \citep{Edwards2015-tn,Vasaikar2019-ob} which includes tumors of any stage; 
    \item DACHS (Darmkrebs: Chancen der Verhütung durch Screening, Southwest Germany) \citep{Hoffmeister2015-vz,Brenner2011-ho}, a large population-based case-control and patient cohort study on CRC, including samples of patients with stages I-IV from different laboratories in southwestern Germany coordinated by the German Cancer Research Center (Heidelberg, Germany);
    \item The DUSSEL (DUSSELdorf, Germany) cohort, a case series of CRC tumors resected with curative intent and collected at the Marien-Hospital in Duesseldorf, Germany, between January 1990 and December 1995 \citep{Grabsch2006-bm};
    \item Epi700 (Belfast, N. Ireland, UK) \citep{Gray2017-bf,Gray2017-oe}, a population-based cohort of stage II and III colon cancers treated by surgical resection between 2003 and 2008;
    \item MECC (Molecular Epidemiology of Colorectal Cancer, Israel) \citep{Shulman2018-mu}, a population-based case-control study in northern Israel; 
    \item The MUNICH (Munich, Germany) CRC series, a case series collected at the Technical University of Munich in Germany.
    \item The NLCS (Netherlands Cohort Study, The Netherlands) \citep{Van_den_Brandt1990-us,Offermans2022-ll} cohort, which contains tissue samples obtained from patients with any tumor stage as part of the Rainbow-TMA consortium study;
    \item QUASAR, the “Quick and Simple and Reliable” trial (Yorkshire, UK) investigating survival benefit of adjuvant chemotherapy in patients from the United Kingdom with mostly stage II tumors \citep{Quirke2007-kf,QUASAR_Collaborative_Group2007-th};
    \item The public repository “The Cancer Genome Atlas”, TCGA (publicly available at \href{https://portal.gdc.cancer.gov/}{https://portal.gdc.cancer.gov/}, USA) \citep{Cancer_Genome_Atlas_Network2012-mk,Isella2014-wf} which includes tumors of any stage; 
    \item The YCR-BCIP (Yorkshire Cancer Research Bowel Cancer Improvement Programme, Yorkshire, United Kingdom [UK]), a population-based register of bowel cancer patients in Yorkshire, UK \citep{West2021-mi,Taylor2019-av}.
\end{enumerate}

Detailed clinicopathological variables are shown in \cref{tab:cohorts}. In all cohorts, formalin-fixed paraffin-embedded (FFPE) tissue was used. Slides have been scanned at their respective centers. For each patient, either an MSI status or a dMMR status, obtained by PCR or IHC, respectively, is available. Although MSI status and dMMR status are not fully concordant \citep{West2021-mi}, they are used interchangeably in clinical routine and grouped as a single category in this study. \textit{KRAS} and \textit{BRAF} mutational status are available for the cohorts DACHS, Epi700, NLCS, QUASAR, and TCGA. 

\subsection{Experimental setup and implementation details}

We performed all experiments using five-fold cross-validation with in-domain validation and testing. In this cross-validation variant, in-domain validation and test set are split off the full data set, leaving three folds for training (\cref{fig:cross_val}). By also cycling the in-domain test set through the complete dataset, we evaluated our model on more representative test sets than when fixing one smaller set for the dataset. During training, the validation set was used to determine the best model, which was consecutively evaluated on the test set. We further evaluated our models on external cohorts outside the dataset used for cross-validation for out-of-domain testing.

The transformer models were trained with the AdamW \citep{Loshchilov2017-ur} optimizer using weight decay and learning rate both of $2*10^{-5}$ . All models were trained for eight epochs with batch size of one for two reasons: first, the sequences of embeddings had different lengths due to the variable number of tiles per WSI and could thus not be stacked to mini-batches of equal length inputs; second, limits in GPU capacity (32GB) because of the quadratic complexity of the self-attention mechanism and the large number of tiles per slide (up to 12,000, \cref{fig:tile_distribution}). To account for the varying cohort size, we evaluated the models every 500 iterations for runs on single cohorts, and every 1000 iterations for runs on multiple cohorts. 

For comparison, we implemented the attention-based MIL approach from \citet{Ilse2018-it}, referred to as AttentionMIL. It provided the best results with Adam \citep{Kingma2014-di} optimizer, $1*10^{-2}$ as weight decay value, along with the fit-one-cycle learning rate scheduling policy \citep{Smith2019-js} with a maximum learning rate of $1*10^{-4}$, trained over 32 epochs, and the first 25\% of the cycle with increasing learning rate. 

\subsection{Statistics and endpoints}

We used the area under the receiver operator curve (AUROC) as our main evaluation metric. Since our data is naturally highly imbalanced with respect to the target variables MSI, \textit{BRAF}, and \textit{KRAS} (\cref{fig:model}d-e), we further used the area under the precision-recall curve (AUPRC) as a metric as this metric accounts better for class imbalances than the AUROC metric. The precision-recall curve relates the recall or specificity, i.e., the ratio of correctly positive predicted samples to all positive samples, to the precision, i.e., the ratio of correctly positive predicted samples to all positive predicted samples. For every experiment, we reported the mean and the standard deviation of respective five-fold cross-validations (\cref{fig:cross_val}). We split the dataset into patient-wise training, validation, and internal test sets stratified by the target label, thus ensuring that every patient can only occur in one of these sets. The external test sets always consisted of different cohorts to better quantify the generalization properties of our algorithms. 

\subsection{Visualization and explainability}

The final prediction is retrieved via the class token that is attached to the input sequence. To visualize the contribution of each input patch, we employed attention rollout as introduced by \citet{Abnar2020-yr}. To obtain the attention at the class token in the final layer, the attention maps of the preceding layers are multiplied recursively. Attention rollout thus quantifies to which extent each patch contributes to the final prediction in the class score. Additionally, we visualized the attention scores for each head in the transformer by taking the class token's self-attention, i.e., the query and key product. All presented attention scores were normalized to the range $[0, 1]$ and clamped to the lower and higher 5\%-quantiles, respectively, for better visual interpretability. 

To visualize whether a patch contributed towards a positive or negative classification outcome, we fed the patches one-by-one through the transformer and visualized the resulting classification scores of the model. These scores were naturally in the range $[0, 1]$ and can thus be directly visualized without further normalizing or clamping of values. 

\subsection{Code availability}
All source codes are available under an open-source license at \href{https://github.com/KatherLab/marugoto/tree/transformer}{https://github.com/KatherLab/marugoto/tree/transformer}. 

\section{Results}

\subsection{A fully transformer-based MSI prediction outperforms the state-of-the-art}

\begin{figure}[t]
    \centering
    \includegraphics[width=\textwidth]{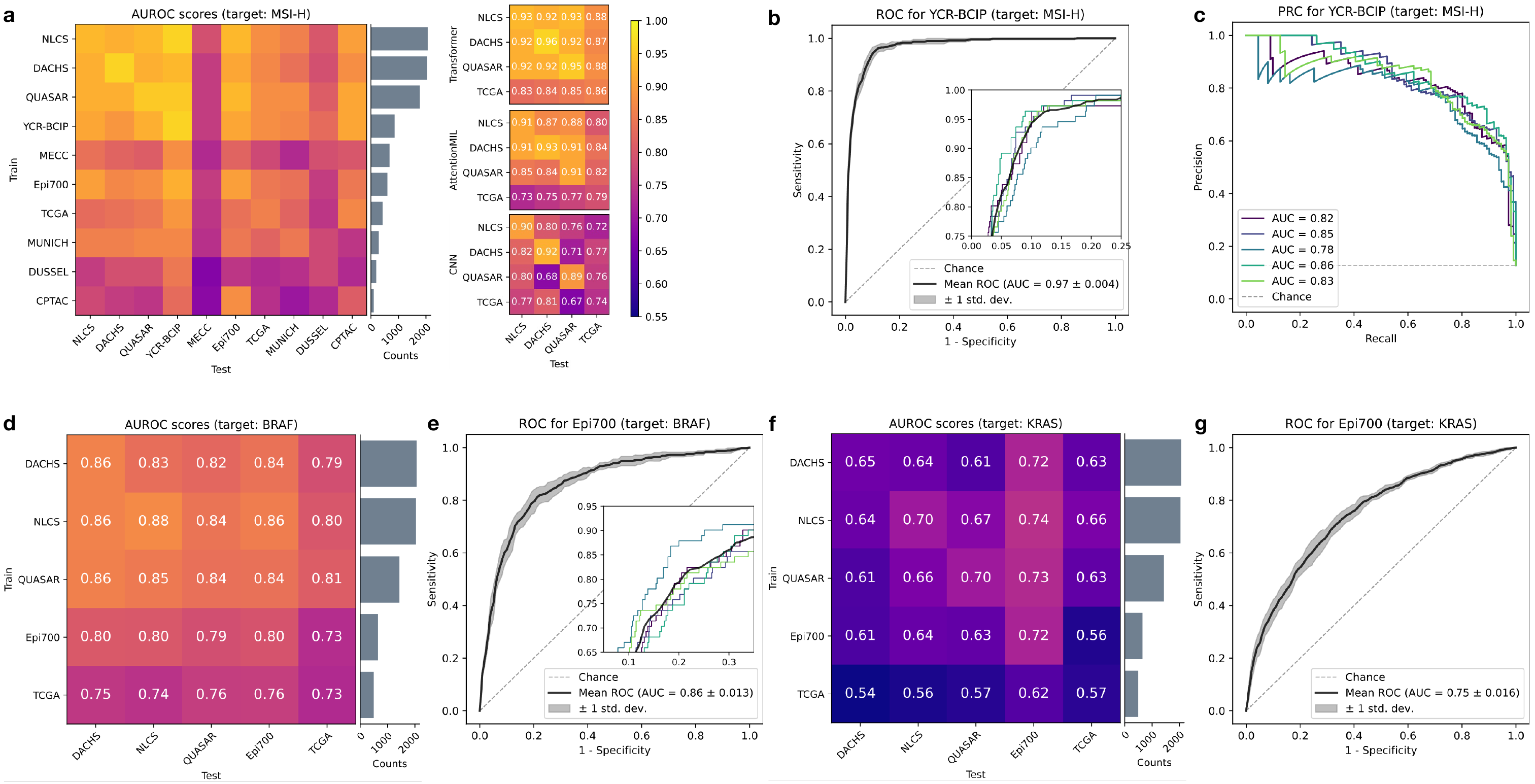}
    \caption{\textbf{Evaluation of the prediction performance for the biomarkers MSI, BRAF, and KRAS in single-cohort and large-scale multi-centric experiments.} Experimental results for MSI-H (a-c), \textit{BRAF} (d,e), and \textit{KRAS} (f,g) prediction: a) AUROC scores for single cohort experiments for all CRC cohorts ordered by size of the cohort. Each row shows the test performance of training on one cohort with the in-domain test results in the diagonal. Results for our transformer approach, AttentionMIL, and CNN approach (results taken from Echle et al.) are visualized separately. Note that compared to AttentionMIL and CNN, our transformer not only shows higher overall prediction accuracy but also better model generalizability, demonstrated by a smaller gap between internal and external testing cohorts. b) Receiver operator curve (ROC) for the model trained on four large cohorts DACHS, QUASAR, NLCS, and TCGA, tested on YCR-BCIP. c) Precision recall curve (PRC) for the model trained on four large cohorts DACHS, QUASAR, NLCS, and TCGA, tested on YCR-BCIP. d) AUROC scores for single cohort experiments. e) ROC for the model trained on four large cohorts DACHS, QUASAR, NLCS, TCGA and tested on Epi700. f) AUROC scores for single cohort experiments g) ROC for the model trained on four large cohorts DACHS, QUASAR, NLCS, TCGA and tested on Epi700. All values represent the mean of 5-fold cross-validation. Raw data for heatmap in a) in \cref{tab:single_cohort}. }
    \label{fig:results}
\end{figure}

We tested our pipeline on MSI prediction in 10 large cohorts of CRC patients in two ways: First, we trained the model on a single cohort and tested it on a held-out test set (in-domain) and on all other cohorts (external). We found that large cohorts, e.g., DACHS, QUASAR, or NLCS, achieved in-domain AUROCs close to 0.95 (\cref{fig:results}a). The model also achieved high performance close to 0.9 AUROC for early-onset CRC, i.e. CRC in patients younger than 50 years (\cref{fig:under50}). We compared this performance to the work by \citet{Echle2020-ot} which updated the CNN-based feature extractor during training and used mean pooling as their patch aggregation function. Our approach outperformed the CNN-based approach on all four cohorts with the largest improvement on TCGA, increasing their results from 0.74 to 0.86 AUROC. Further, we also evaluated AttentionMIL by \citet{Ilse2018-it} with CTransPath as feature extractor yielding higher performance than the CNN-based approach on the large cohorts, but partly lower results on the external validation trained on the smaller cohort TCGA.

Second, we mirrored the experimental setup of \citet{Echle2020-ot} and trained AttentionMIL and our fully transformer-based model on four large cohorts (DACHS, NLCS, QUASAR, and TCGA) and evaluated them on the external cohort YCR-BCIP. We used CTransPath as a feature extractor for both, AttentionMIL and our transformer-based model. The CNN-based approach from Echle et al. achieved an AUROC of 0.96, AttentionMIL yielded an AUROC of 0.96, and the fully transformer-based approach performed slightly better with an AUROC of 0.97 (\cref{fig:results}b). In particular, we obtained a sensitivity of 0.97 with a negative predictive value of 0.99 (\cref{fig:cm_mean}). Moreover, a high mean AUPRC score of 0.83 showed that the transformer-based model achieved high sensitivity with relatively high precision despite a high class imbalance of 12.9\% MSI-high samples on average across all cohorts (\cref{fig:results}c).

The classical patch-based approach by \citet{Echle2020-ot} suffered from severe losses in performance upon external testing. The largest performance drop in the AUROC of 0.21 was observed by a model trained on the DACHS and tested on the QUASAR cohort. Our transformer model, however, reduced the performance loss for external testing to a maximum of 0.08 for training on the NLCS and testing on the TCGA cohort. In addition, AttentionMIL trained with the same transformer-based feature extractor also demonstrated better generalization capabilities compared to the classical patch-based approach with mean pooling. This suggests that self-supervised pretraining on histology data contributes positively towards better generalization.

In summary, these results show that a fully transformer-based approach yields a higher performance for biomarker prediction both on large cohorts (DACHS, QUASAR, and NLCS) as well as on smaller cohorts (TCGA). Perhaps more importantly from a clinical perspective, the transformer-based approach resulted in a better generalization performance and more reliable results, as the deviation between the external cohorts was smaller. We published all trained models for re-use, and further fine-tuning if needed.

\subsection{The fully transformer-based model predicts multiple biomarkers in CRC}

Next, we investigated whether the fully transformer-based model yields a similar high performance in other biomarker prediction tasks. Following the experimental setup for MSI prediction, we trained the model first on single cohorts evaluated on all other external cohorts and second on one fully merged multi-center cohort excluding only one cohort from the training set to constitute an external test set. In clinical routine workup for CRC, the biomarkers \textit{BRAF} and \textit{KRAS} are determined in addition to MSI. We tested whether and how well these were predictable on the DACHS, QUASAR, NLCS, TCGA, and Epi700 cohorts, where the Epi700 cohort served as an external test set in the multi-center run. 

In the case of our larger cohorts, namely DACHS, QUASAR, and NLCS, single cohort training was already capable of achieving good results, with AUROCs of 0.86, 0.84, and 0.88, respectively (\cref{fig:results}d). The smaller cohorts achieved poorer results with wider standard deviations in AUROC across the five folds (0.1 and 0.08 for TCGA and Epi700 respectively, as compared to 0.02-0.03 for the larger cohorts). Nonetheless, the AUROC for the in-domain test using TCGA outperformed previous approaches with AUROCs of 0.57 \citep{Fu2020-tw} or 0.66 \citep{Kather2020-kr}, and performed on par with more recent transformer-based methods with an AUROC of 0.73 \citep{Reisenbuchler2022-dk}. The large multi-centric cohort yielded an AUROC of 0.86 (\cref{fig:results}e). Furthermore, we observed that the generalization gap from the internal test set to external cohorts was consistently small with the largest internal-to-external gap observed for the smallest cohort, TCGA. This was also the case in multi-centric evaluation, where the performance did not decrease from the internal to the external test cohort. 

We observed similar results regarding the generalization when investigating \textit{KRAS} as a target (\cref{fig:results}f-g), with an AUROC of 0.75 when trained on the multi-centric cohort, and AUROCs ranging from 0.57 to 0.72 for single cohorts. This is in line with state-of-the-art results \citep{Kather2020-kr,Fu2020-tw,Reisenbuchler2022-dk}. While DL-based prediction performance for \textit{KRAS} is still relatively low compared to MSI or \textit{BRAF}, the results show that performance increases with the number of patients in the training cohort. 

\begin{figure}[thp]
    \centering
    \includegraphics[width=\textwidth]{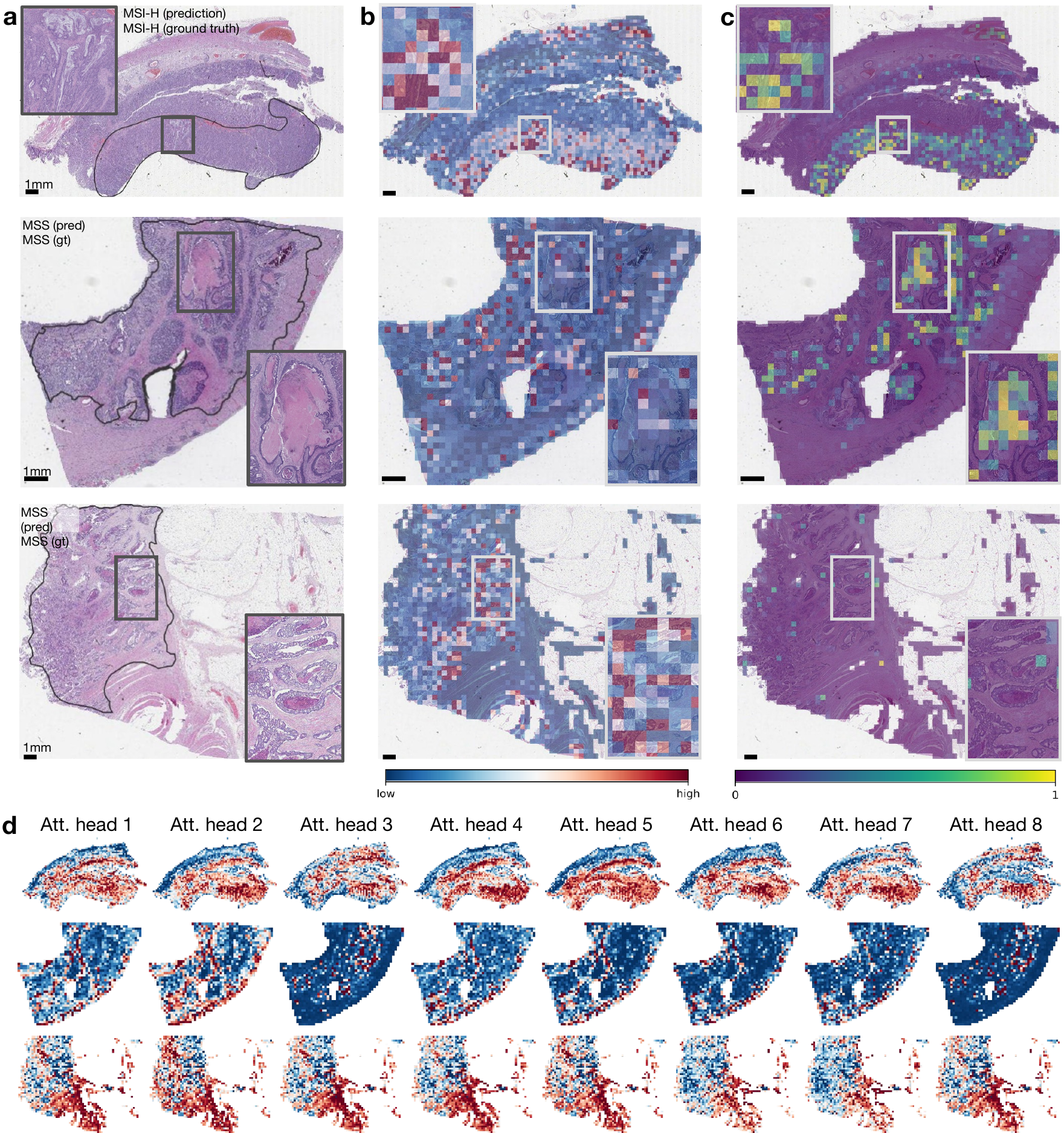}
    \caption{\textbf{Attention and class score visualization for better model interpretability.}  a) Resection specimen from the external cohort YCR-BCIP. The three depicted slides are the same as in \citet{Echle2020-ot}. Tumor regions are outlined in black. b) Attention rollout per patch for our trained transformer-based feature aggregation model. Large values (red) signify a high contribution to the model’s prediction, smal values (blue) a low contribution. c) MSI classification scores per patch, where MSI-high is the positive class (scores larger then 0.5) and MS-stable is the negative class (scores lower than 0.5). d) The attention heatmaps from the all eight heads of the last layer. The methods used for visualization are describesd in the methods section. We used the model weights from the best performing fold of the multi-centric training on DACHS, NLCS, QUASAR, and TCGA.}
    \label{fig:attention}
\end{figure}

Overall, these findings show that our model can predict multiple biomarkers relevant for routine diagnostics in CRC while highlighting the importance of large training cohorts to reach clinically relevant performance.

\subsection{Fully transformer-based workflows are explainable}

Ideally, DL-based biomarker predictions should be explainable to domain experts. To this end, we visualized how much each patch contributed to the final classification via attention rollout as well as whether it contributes towards a positive or negative classification (\cref{fig:attention}a-c). 

For better comparability, we used the same WSIs from the external cohort YCR-BCIP as had been used in a previous study \citep{Echle2020-ot} for these visualizations (\cref{fig:attention}a). For all three cases, the majority of highly-contributing patches originate from tumor regions. In the MSI-high case, the mucinous region that is morphologically linked to MSI is correctly identified as important by high scores in the attention rollout as well as the patch-level classification. The MS-stable case in the second row of \cref{fig:attention}a-c attributes high contributions to the model’s prediction to carcinoma regions. At the same time these patches all receive low classification scores yielding the correct classification result. Similarly, the second MS-stable case in the third row had highly contributing scores in the tumor region while having only low classification scores for all patches. Further tissue details that are morphologically related to MSI, such as solid growth patter, poor differentiation, or lymphocytes (\cref{fig:annotations}) are highlighted in the attention heads of the last layer together with healthy tissue structures, such as the colon wall including muscle tissue or vessels (\cref{fig:attention}d).

We observed that our model attributes a high level of attention not only to tissue regions, such as necrotic regions, but also larger blood vessels. Notably, such structures are not considered morphologically related to MSI from a pathologist's perspective. Since these regions are particularly intensely stained by eosin, i.e., very pink or red, we suspect that the high color intensities lead to these responses. For all three cases, vessels in healthy tissue got higher attention scores than other healthy tissue regions but no high class scores, and therefore did not influence the MSI classification positively. However, the necrotic region highlighted in the second row of \cref{fig:attention} had high class scores similar to those seen in the model by \citet{Echle2020-ot}.

These examples showed that the model learns concepts relevant to MSI-high prediction and thus possess a high degree of explainability. Visualizing the attention rollout together with the classification scores demonstrates that relevant regions can receive high attention scores while the model can learn to ignore non-relevant regions or give them low class scores. 

\subsection{Fully transformer-based workflows are more data efficient}

\begin{figure}[t]
    \centering
    \includegraphics[width=\textwidth]{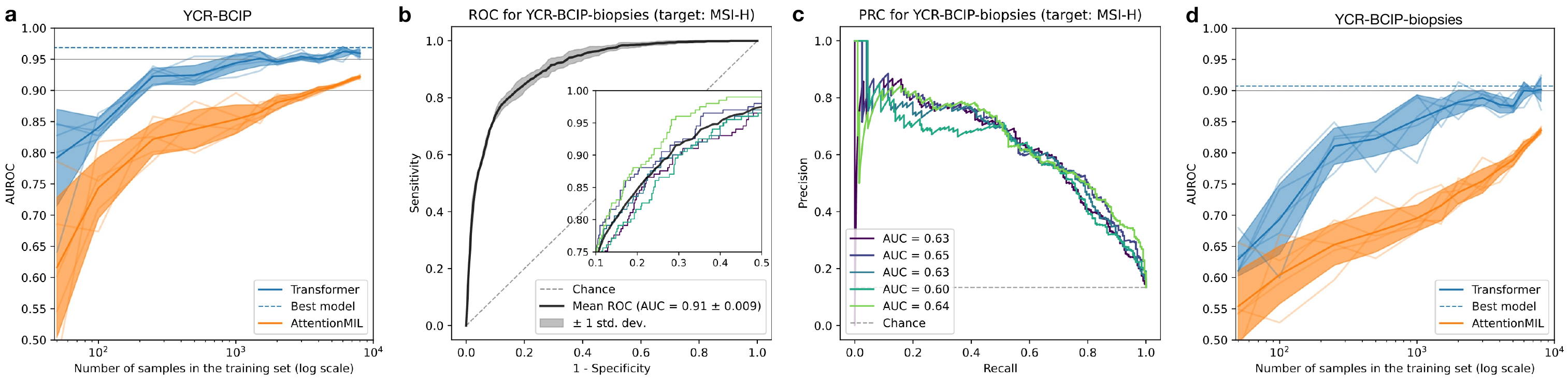}
    \caption{\textbf{Analysis of data efficiency and model generalization to biopsies.} a) AUROC scores on YCR-BCIP depending on the number of patients available for training. The samples were randomly drawn from the merged cohorts DACHS, NLCS, QUASAR, and TCGA. b) ROC for testing our model on YCR-BCIP-biopsies, trained on resections from DACHS, NLCS, QUASAR, and TCGA. c) PRC for testing our model on YCR-BCIP-biopsies, trained on resections from DACHS, NLCS, QUASAR, and TCGA. d) Analogue analysis to a) tested on YCR-BCIP-biopsies.}
    \label{fig:results_biopsies}
\end{figure}

A longstanding problem in computational pathology is to determine the number of samples required for a given prediction task. This is primarily due to two reasons. First, it is unclear what the minimum required sample size is, and second, it is unclear if adding more samples improves performance, and if so, up to what point. To investigate this, we varied the number of patients in the training set and analyzed its impact on the test performance. Specifically, we merged all cohorts except for an external validation cohort, YCR-BCIP, resulting in a training set with 8181 patients from nine cohorts. We trained models using a fixed number of epochs and randomly sampled patients from the training set. All experiments were repeated five times and we reported the means and standard deviations of the results.

Our fully-transformer based model architecture achieved a mean testing AUROC value above 0.9 with 250 patients (in particular, an AUROC value of 0.92), while the AttentionMIL model exceeded an AUROC of 0.9 only with 4000 patients (\cref{fig:results_biopsies}a). In a similar vein, our model surpassed the 0.95 mean testing AUROC with already 1500 patients, while AttentionMIL did not reach this performance. Hence, the transformer-based aggregation module helped the model to learn from data in a more efficient way than the attention mechanism. This may be due to the attention mechanism not contextualizing information from all input patches. Of note, above 1000 patients, the performance of the transformer-based model seems to slowly saturate, while the Attention mechanism continues to increase in performance with more patients but on a lower level. 

Our fully transformer-based approach yielded high performance with a small sample size. Compared to an AttentionMIL-based approach, our approach is more data efficient in the regime of small numbers of patients. Looking at larger training numbers, we observed that performance increase is directly proportional to the number of patients for both approaches, but the fully transformer-based approach reaches equivalent performance already with much smaller datasets.

\subsection{Fully transformer-based workflows result in clinical-grade performance on biopsies}

Virtually all previous studies on biomarker prediction in CRC were performed using surgical resection slides. For this reason, commercially available MSI detection algorithms are intended to be used only with resection slides. However, recent clinical evidence shows that MSI-positive CRC patients require immunotherapy prior to surgery \citep{Cercek2022-na,Chalabi2022-tp} and hence need to be tested for MSI on biopsy material. We addressed this problem by training our model on resections from the cohorts, DACHS, QUASAR, NLCS, and TCGA, and evaluating it on biopsies from 1,502 patients with CRC of the YCIP-BCIP. 

Our model yielded a mean AUROC score of 0.91 when externally validated on biopsies (\cref{fig:results_biopsies}b). We outperformed existing approaches (0.78 by  \citep{Echle2020-ot}) by far and achieved clinical-grade performance on biopsies after model training on resection specimen (\cref{fig:cm_mean}). However, the mean AUPRC score of 0.63 however was lower than that of the external cohort YCR-BCIP for resections (0.83) (\cref{fig:results_biopsies}c). Hence, choosing a classification threshold with high sensitivity, the ratio of correctly MSI positive predicted cases from all positive predicted cases was lower on biopsies compared to resections. 

\begin{figure}[ht]
    \centering
    \includegraphics[width=\textwidth]{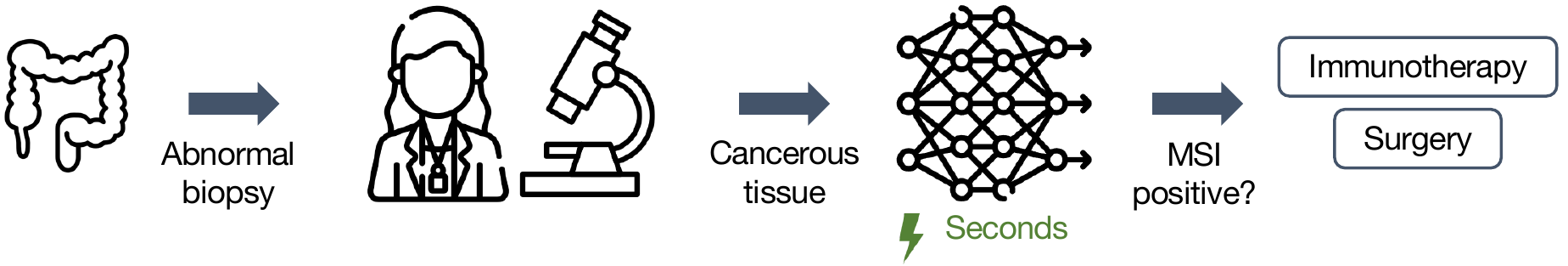}
    \caption{\textbf{Envisioned clinical workflow for the proposed MSI-high classifier on biopsies.} This assumes the system reaches a sufficient performance in additional external validation and is approved as a medical device. This workflow would only apply to non-metastatic disease. Neoadjuvant immunotherapy is not yet recommended by medical guidelines but is backed up by Phase-II clinical trials. Not shown: tissue preprocessing and scanning pipelines, and confirmatory tests of MSI after a positive deep learning-based pre-screening. }
    \label{fig:biopsy_workflow}
\end{figure}

Our intended clinical use of this workflow is as follows (\cref{fig:biopsy_workflow}): First, a patient attends a clinic either with suspected CRC or for routine CRC screening. A colonoscopy shows a suspicious tumor, which is evaluated histologically and found to be an adenocarcinoma. In many countries, this biopsy will then be tested for MSI/MMR status and \textit{BRAF} and/or \textit{RAS} mutation status. In practice, these procedures may take several days to even weeks. However, in low-or-middle-income countries, this might not happen at all. Based on the MSI, \textit{BRAF}, and \textit{RAS} status, the most suitable treatment approach will be chosen for the patient. For example, in patients with early (non-metastatic) CRC, the presence of MSI could qualify a patient for neoadjuvant immunotherapy followed by surgery with curative intent. Similarly, in the metastatic disease setting, the presence of MSI in the biopsy tissue would qualify a patient for palliative immunotherapy. Our algorithm could therefore speed up the step between taking the biopsy and the molecular determination of MSI-high status, thus enabling an earlier treatment with immunotherapy if indicated. 

In summary, to the best of our knowledge, we developed the first DL-based MSI-high predictor for biopsies that achieves clinical-grade performance. In particular, this high performance was also observed for external tests and could therefore improve clinical routine and speed up treatment decisions. 

\section{Discussion}

The rollout of precision oncology to colorectal cancer patients promises gains in life expectancy \citep{Hendricks2020-it}. Unfortunately, however, its implementation still remains slow and patchy. One for this is that precision oncology biomarkers are complex, costly and require intricate instrumentation and expertise. DL is emerging as a possible solution for this problem \citep{Shmatko2022-zi,Bera2019-rg}. DL can extract biomarker information directly from routinely available material, thereby potentially providing cost savings \citep{Kather2019-zm}. Using DL-based analysis of histopathology slides to extract biomarkers for oncology has become a common approach in the research setting in 2018 \citep{Coudray2018-qo}. In turn, this has recently led to regulatory approval of multiple algorithms for clinical use. Some of these examples include a breast cancer survival prediction algorithm by Paige (New York, NY, USA), a method to predict survival in CRC by DoMore Diagnostics (Oslo, Norway), a method to predict MSI status in CRC by Owkin (Paris, France and New York, NY, USA), amongst others. However, existing DL biomarkers have some key limitations: it is debated whether or not their performance is sufficient for large-scale use, they do not necessarily generalize to any patient population, and finally, they are not approved for use on biopsy material, as the application of DL algorithms to biopsies typically results in much lower performance compared to application to surgical specimens \citep{Echle2020-ot}. 

A key reason for the limited performance of existing DL systems could be the fundamental limitations of the technology employed. Most studies between 2018 and 2020 used convolutional neural networks (CNNs) as their DL backbone \citep{Ghaffari_Laleh2022-ic}, using publicly available information. Commercial products in the DL biomarker space are based on the same technology \citep{Campanella2019-gb,Svrcek2022-ck,Kleppe2022-fx}. However, a new class of neural networks has recently started to replace CNNs: transformers. Originating from the field of natural language processing, transformers are a powerful tool to process sequences and leverage the potential of large amounts of data. Also in computer vision, transformers yield a higher accuracy for image classification in non-medical tasks \citep{Dosovitskiy2020-gz,Liu2021-up}, are more robust to distortions in the input data \citep{Ghaffari_Laleh2022-sr} and provide more detailed explainability \citep{Chen2022-ej}. These advantages of transformers compared to CNNs have the potential to translate into more accurate and more generalizable clinical biomarkers, but there is currently no evidence to support this.

In the present study, we developed the first fully transformer-based approach for biomarker prediction on whole-slide images of H\&E-stained CRC tissue sections. Our model consists of a transformer-based feature extractor that was pretrained on histopathology images and a transformer-based aggregation module. In contrast to the state-of-the-art attention-based MIL approaches, the contribution of each patch was not only determined according to its feature embeddings but also contextualized with the feature embeddings of all other patches in the WSI via self-attention layers. Further, we presented the first large-scale evaluation of transformers in biomarker prediction on WSIs. We demonstrated that transformer-based approaches learned better from small amounts of data and were therefore more data-efficient than attention-based MIL approaches. At the same time, the performance increased proportionally with the number of training samples. Even though the performance seemed to plateau for MSI prediction, this suggests that larger training cohorts could lead to higher performance approaching clinical application, also for more challenging tasks such as the prediction of the \textit{BRAF} and \textit{KRAS} mutational status. Our large-scale evaluation also showed that MIL and in particular transformer-based approaches generalize much better than the existing CNN approaches. We proved this by training the model on single cohorts and testing the generalization on all other cohorts. These experiments showed that the transformer-based approach reduced the drop in AUROC to under 0.08, while the CNN-based approach dropped by more than 0.21 on some cohorts. Most importantly, our approach trained on resections did not only generalize well to external cohorts with resections but also to biopsies with a clinical-grade performance of 0.91 AUROC.
A caveat of our experiments is that the ground truth is not perfect. Potentially, the DL model is performing better than stated in the paper because dMMR and MSI only agree around 92\% of the time \citep{West2021-mi,NICEguideline}. Also, there are POLD1 and POLE mutant cases with a high mutation burden that are not identified by IHC, but behave clinically similar to MSI. Now the DL tests are high levels of performance so these small subpopulations may be important. Which are the cases that are picked up by the transforming algorithm that are not picked up by the old methods - this should be investigated in the future. 

Our study still had some limitations: The focus of this study was to investigate the effect of handling the data with fully transformer-based approaches, especially in the context of large-scale multi-institutional data. Therefore, we did not optimize all details and hyperparameters of the architecture. Points for optimization in this direction would be finding a fitting positional encoding and tuning the architecture of the transformer network. Further, we applied Macenko stain normalization in the preprocessing pipeline since this reduced the generalization gap compared to no stain normalization, but other methods in this direction could be investigated to further reduce the generalization drop even further. Additionally, collecting biopsy samples from different hospitals, for multi-cohort training directly on biopsy data could potentially improve the performance of our model on biopsy material. This would also hold for the prediction of \textit{BRAF} mutation status and, in particular, of \textit{RAS} mutation status, where we observed the largest potential for improvement. In both targets, the performance was higher on the larger cohorts with around 2,000 patients and increased dramatically by training on multiple cohorts.

In summary, to the best of our knowledge, we presented the first fully transformer-based model to predict MSI on WSI from CRC with an AUROC of 0.97 on resections and 0.91 on biopsies, both obtained with external validation. Our model generalized better to unseen cohorts and was more data-efficient compared to existing state-of-the-art MIL or CNN approaches. By publishing all trained models, we enable researchers and clinicians to apply the automated MSI prediction tool in clinical practice, which we expect to bring the field of DL-based biomarkers a step closer to integration in the clinical workflow.

\newpage

\section*{Additional Information}

\subsection*{Disclosures}

JNK reports consulting services for Owkin, France, Panakeia, UK, and DoMore Diagnostics, Norway and has received honoraria for lectures by MSD, Eisai, and Fresenius. NW has received fees for advisory board activities with BMS, Astellas, and Amgen, not related to this study. NW has received fees for advisory board activities with BMS, Astellas, and Amgen, not related to this study. PQ has received fees for advisory board activities with Roche and AMGEN and research funding from Roche through an Innovate UK National Pathology Imaging Consortium grant. HIG has received fees for advisory board activities by AstraZeneca and BMS, not related to this study. MST is a scientific advisor to Mindpeak and Sonrai Analytics, and has received honoraria recently from BMS, MSD, Roche, Sanofi and Incyte. He has received grant support from Phillips, Roche, MSD and Akoya. None of these disclosures are related to this work. No other potential disclosures are reported by any of the authors. 

\subsection*{Author contributions}

SJW, MB, TP, and JNK designed the concept of the study, MB, GPV supported with domain-related advice; DR, SJW, MB, and TP developed and evaluated a preliminary study, where DR implemented the methods; SJW, JMN prepared the data for this study, SJW implemented the methods and ran all experiments and evaluations; NW, PQ, HIG, PvdB, GGAH, SDR, KH, CG, MPE, ME, MB, DH, CR, TY, RL, JCAJ, KO, WM, RG, SBG, JKG, GR, JDB, DS, JJ, ML, MST, HB, and MH provided clinical and histopathological data; all authors provided clinical expertise and contributed to the interpretation of the results. SJW wrote the manuscript with JNK, and input from all other authors.

\subsection*{Acknowledgments}

DACHS study: The authors thank the hospitals recruiting patients for the DACHS study and the cooperating pathology institutes. We thank the National Center for Tumor Diseases (NCT) Tissue Bank, Heidelberg, Germany, for managing, archiving, and processing tissue samples in the DACHS study.

We acknowledge the support of the Rainbow-TMA Consortium, especially the project group: PA van den Brandt, A zur Hausen, HI Grabsch, M van Engeland, LJ Schouten, J Beckervordersandforth (Maastricht University Medical Center, Maastricht, The Netherlands); PHM Peeters, PJ van Diest, HB Bueno de Mesquita (University Medical Center Utrecht, Utrecht, The Netherlands); J van Krieken, I Nagtegaal, B Siebers, B Kiemeney (Radboud University Medical Center, Nijmegen, The Netherlands); FJ van Kemenade, C Steegers, D Boomsma, GA Meijer (VU University Medical Center, Amsterdam, The Netherlands); FJ van Kemenade, B Stricker (Erasmus University Medical Center, Rotterdam, The Netherlands); L Overbeek, A Gijsbers (PALGA, the Nationwide Histopathology and Cytopathology Data Network and Archive, Houten, The Netherlands); and Rainbow-TMA collaborating pathologists, among others: A de Bruïne (VieCuri Medical Center, Venlo); JC Beckervordersandforth (Maastricht University Medical Center, Maastricht); J van Krieken, I Nagtegaal (Radboud University Medical Center, Nijmegen); W Timens (University Medical Center Groningen, Groningen); FJ van Kemenade (Erasmus University Medical Center, Rotterdam); MCH Hogenes (Laboratory for Pathology Oost-Nederland, Hengelo); PJ van Diest (University Medical Center Utrecht, Utrecht); RE Kibbelaar (Pathology Friesland, Leeuwarden); AF Hamel (Stichting Samenwerkende Ziekenhuizen Oost-Groningen, Winschoten); ATMG Tiebosch (Martini Hospital, Groningen); C Meijers (Reinier de Graaf Gasthuis/SSDZ, Delft); R Natté (Haga Hospital Leyenburg, The Hague); GA Meijer (VU University Medical Center, Amsterdam); JJTH Roelofs (Academic Medical Center, Amsterdam); RF Hoedemaeker (Pathology Laboratory Pathan, Rotterdam); S Sastrowijoto (Orbis Medical Center, Sittard); M Nap (Atrium Medical Center, Heerlen); HT Shirango (Deventer Hospital, Deventer); H Doornewaard (Gelre Hospital, Apeldoorn); JE Boers (Isala Hospital, Zwolle); JC van der Linden (Jeroen Bosch Hospital, Den Bosch); G Burger (Symbiant Pathology Center, Alkmaar); RW Rouse (Meander Medical Center, Amersfoort); PC de Bruin (St. Antonius Hospital, Nieuwegein); P Drillenburg (Onze Lieve Vrouwe Gasthuis, Amsterdam); C van Krimpen (Kennemer Gasthuis, Haarlem); JF Graadt van Roggen (Diaconessenhuis, Leiden); SAJ Loyson (Bronovo Hospital, The Hague); JD Rupa (Laurentius Hospital, Roermond); H Kliffen (Maasstad Hospital, Rotterdam); HM Hazelbag (Medical Center Haaglanden, The Hague); K Schelfout (Stichting Pathologisch en Cytologisch Laboratorium West-Brabant, Bergen op Zoom); J Stavast (Laboratorium Klinische Pathologie Centraal Brabant, Tilburg); I van Lijnschoten (PAMM Laboratory for Pathology and Medical Microbiology, Eindhoven); and K Duthoi (Amphia Hospital, Breda).

\subsection*{Funding}

SJW and DR are supported by the Helmholtz Association under the joint research school “Munich School for Data Science - MUDS” and SJW is supported by the Add-on Fellowship of the Joachim Herz Foundation. JNK is supported by the German Federal Ministry of Health (DEEP LIVER, ZMVI1-2520DAT111), the Max-Eder-Programme of the German Cancer Aid (grant \#70113864) and the German Academic Exchange Service (SECAI, 57616814). JNK and MH are funded by the German Federal Ministry of Education and Research (PEARL, 01KD2104C). PQ, NW, SD, and GH are supported by Yorkshire Cancer Research grants L386 and L394. PQ, HG, NW, JNK, and SD are supported in part by the National Institute for Health and Care Research (NIHR) Leeds Biomedical Research Centre. The views expressed are those of the author(s) and not necessarily those of the NHS, the NIHR, or the Department of Health and Social Care. PQ is also supported by an NIHR Senior Investigators award. Tumor tissue collection in the NLCS was done in the Rainbow-TMA study, which was financially supported by BBMRI-NL, a Research Infrastructure financed by the Dutch government (NWO 184.021.007 to PvdB). The analyses of MSI, BRAF, and KRAS in the NLCS were funded by The Dutch Cancer Society (KWF 11044 to PvdB). The DACHS study (HB, JCC, and MH) was supported by the German Research Council (BR 1704/6-1, BR 1704/6-3, BR 1704/6-4, CH 117/1-1, HO 5117/2-1, HO 5117/2-2, HE 5998/2-1, HE 5998/2-2, KL 2354/3-1, KL 2354/3-2, RO 2270/8-1, RO 2270/8-2, BR 1704/17-1 and BR 1704/17-2), the Interdisciplinary Research Program of the National Center for Tumor Diseases (NCT; Germany) and the German Federal Ministry of Education and Research (01KH0404, 01ER0814, 01ER0815, 01ER1505A and 01ER1505B). The study was further supported by project funding for the PEARL consortium from the German Federal Ministry of Education and Research (01KD2104A).

\bibliographystyle{unsrtnat}
\bibliography{references}  

\pagebreak
\appendix
\counterwithin{figure}{section}
\counterwithin{table}{section}

\section{Supplementary Materials}

\subsection{Supplementary Figures}

\begin{figure}[H]
    \centering
    \includegraphics[width=0.2\textwidth]{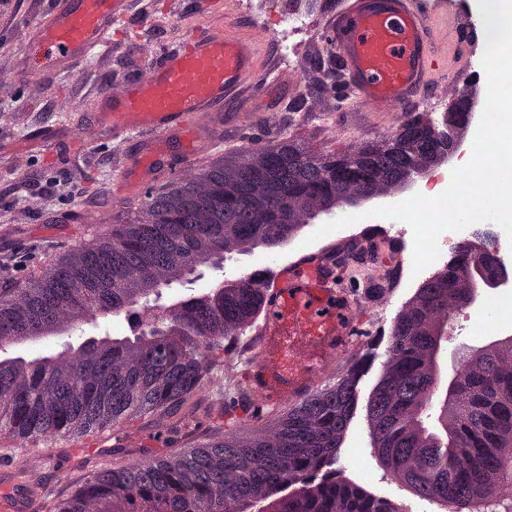}
    \caption{\textbf{Template tile used for stain normalization with Macenko’s method.} The tile was taken from the DACHS cohort.}
    \label{fig:normalization_template}
\end{figure}

\begin{figure}[H]
    \centering
    \includegraphics[width=0.5\textwidth]{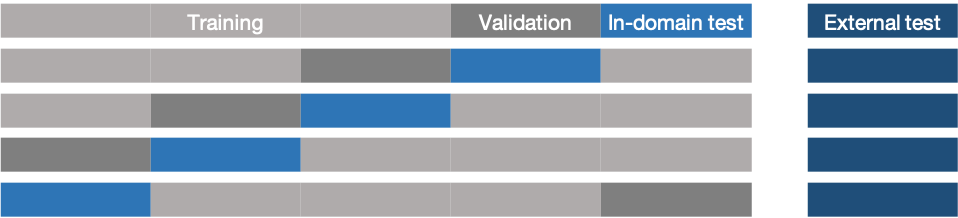}
    \caption{\textbf{Five-fold cross-validation scheme.} Additionally to the standard use of five-fold cross-validation, we also cycle the in-domain test set through the whole dataset to get a more representative test evaluation of the dataset compared to splitting up a small in-domain test set beforehand.}
    \label{fig:cross_val}
\end{figure}

\begin{figure}[H]
    \centering
    \includegraphics[width=0.95\textwidth]{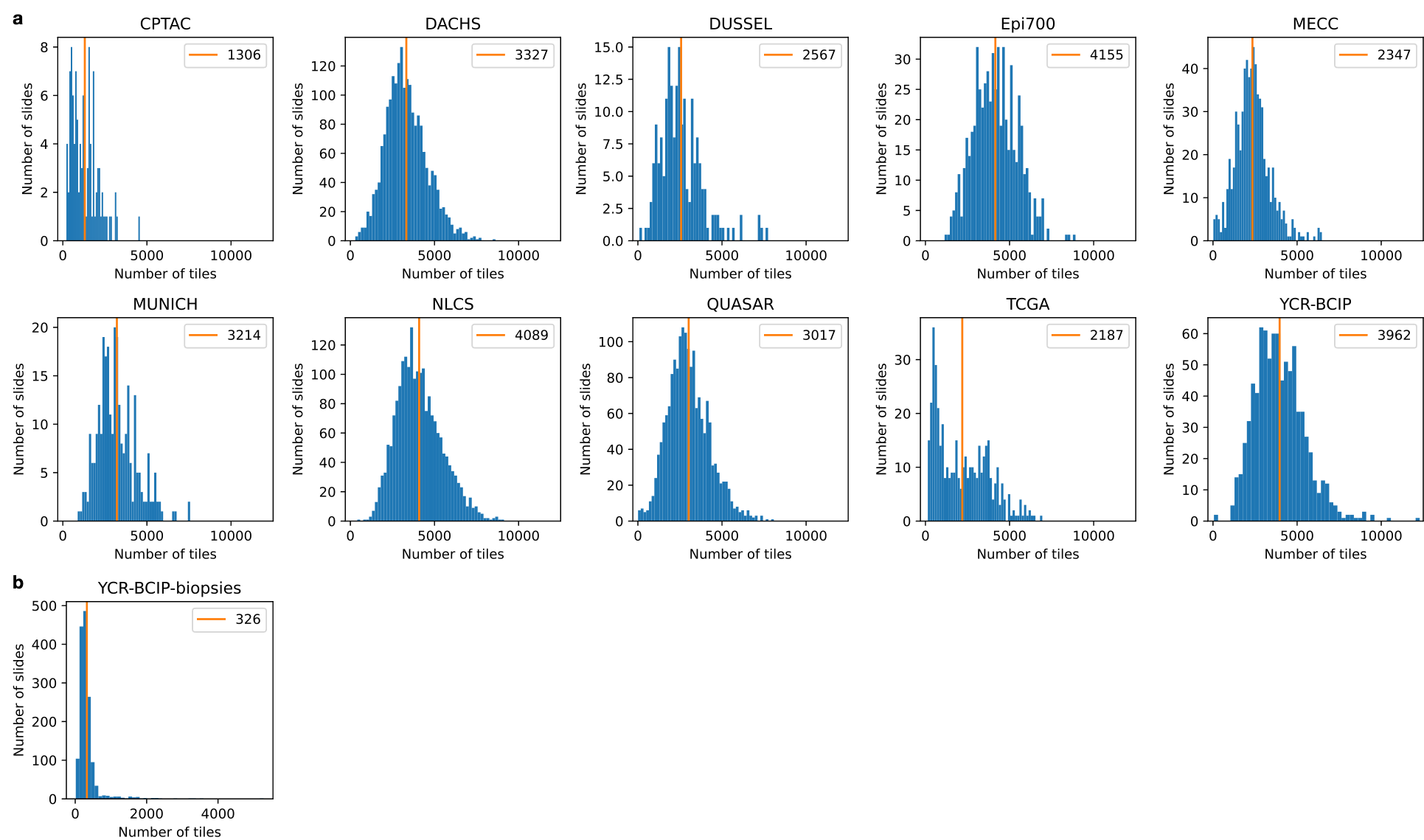}
    \caption{\textbf{Distribution of the number of tiles per slide for all cohorts.} The mean of each distribution is highlighted in orange. a) Histogram of the distribution for all 10 cohorts of resections. b) Histogram of the distribution for the biopsy cohort.}
    \label{fig:tile_distribution}
\end{figure}

\begin{figure}[H]
    \centering
    \includegraphics[width=\textwidth]{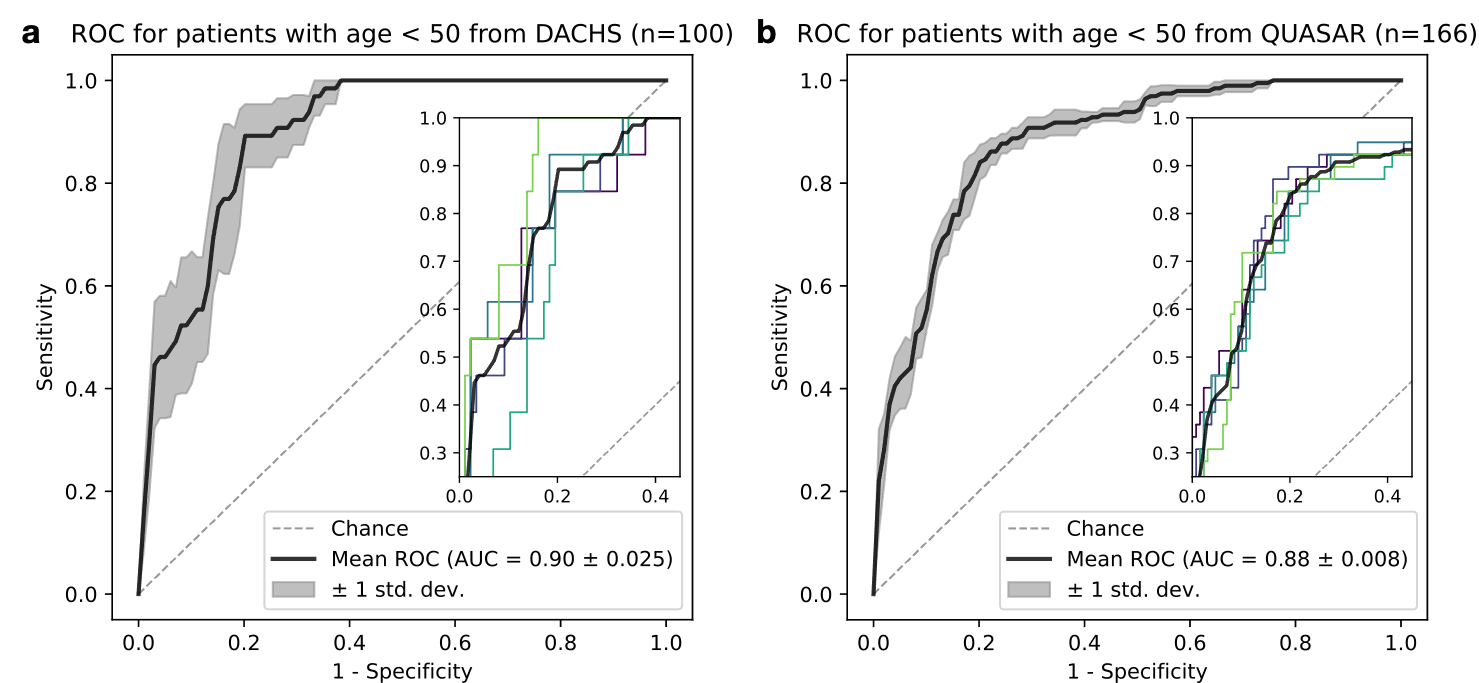}
    \caption{\textbf{Receiver operator curves for early onset cancer.} a) External test on DACHS with 100 patients younger than 50 years, trained on QUASAR. b) External test on QUASAR with 166 patients younger than 50 years, trained on DACHS. Both cohorts were chosen because they contain a sufficient number of patients under 50 in contrast to the other cohorts in this study.}
    \label{fig:under50}
\end{figure}

\begin{figure}[H]
    \centering
    \includegraphics[width=\textwidth]{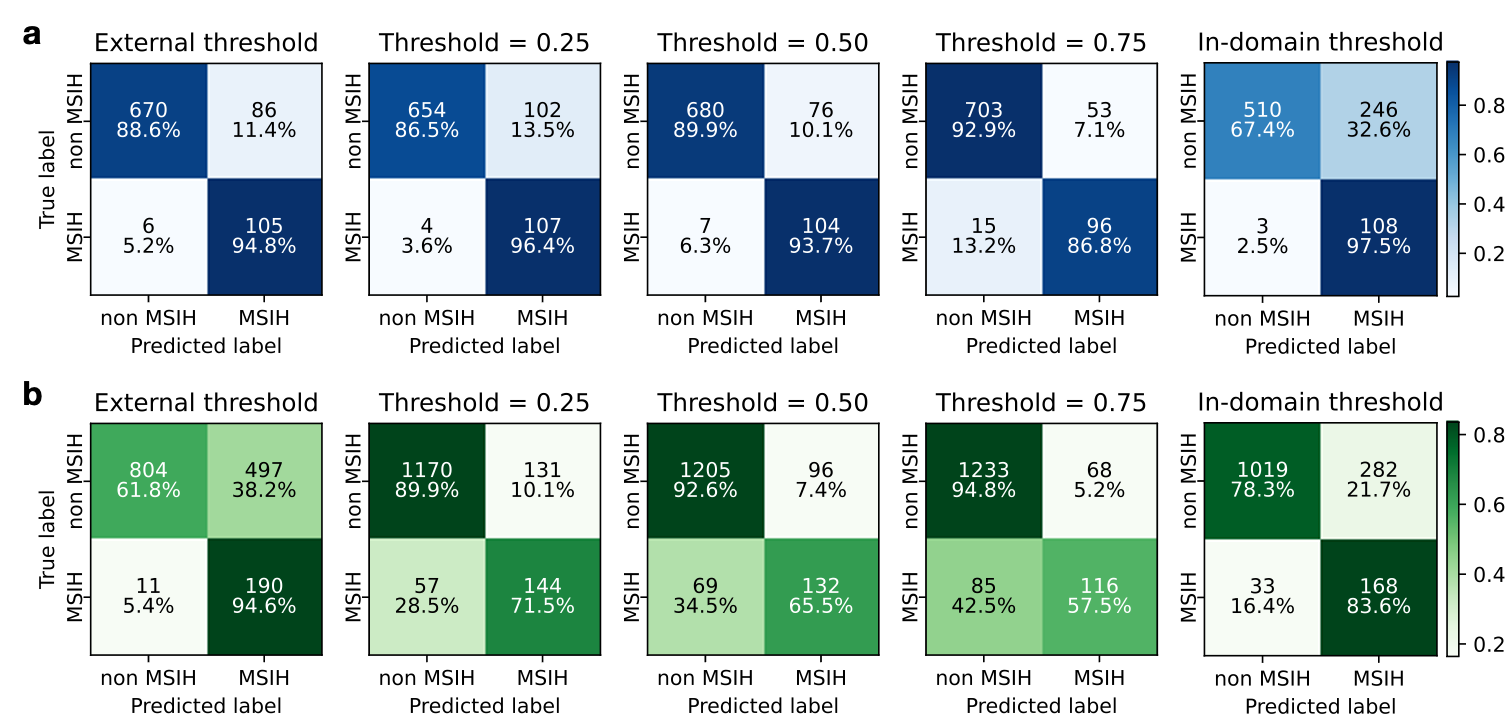}
    \caption{\textbf{Confusion matrix for different classification thresholds.} First column shows the threshold determined on the external tests, such that 0.95 sensitivity is reached. Second to forth column show foxed thresholds (0.25, 0.5, 0.75, respectively). Last column shows the threshold determined on the in-domain test set, such that 0.95 sensitivity is reached. The results show the average of the model trained on the large multi-centric cohort DACHS, NLCS, QUASAR, and TCGA across all five folds. a) Results for the resection cohort YCR-BCIP. b) Results for the biopsy cohort YCR-BCIP-biopsies.}
    \label{fig:cm_mean}
\end{figure}

\begin{figure}[H]
    \centering
    \includegraphics[width=\textwidth]{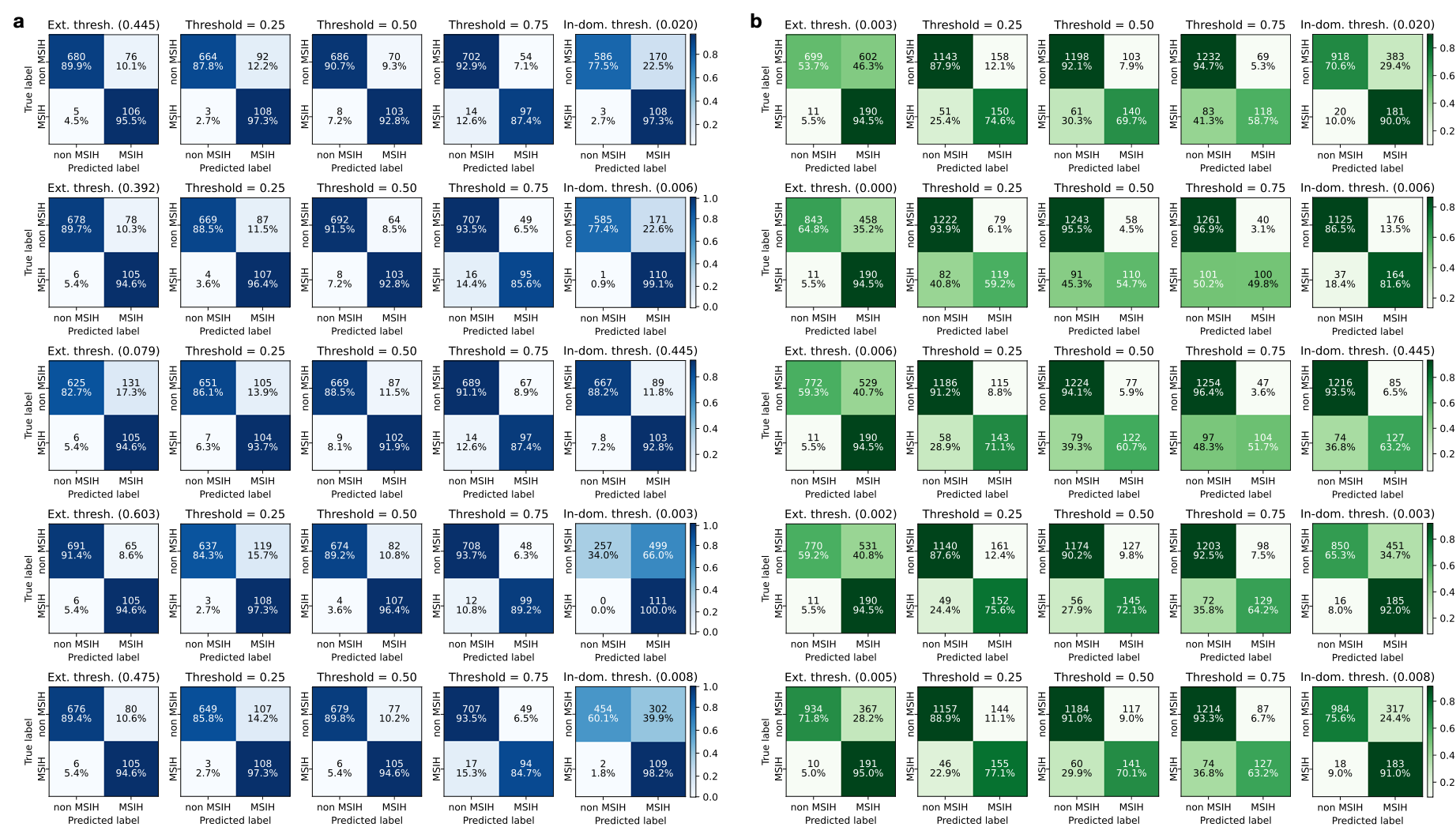}
    \caption{\textbf{Confusion matrix for different classification thresholds and all five folds.} Similar to  \cref{fig:cm_mean}. Here the results of the models of each fold are shown in one row. a) Results for the resection cohort YCR-BCIP. b) Results for the biopsy cohort YCR-BCIP-biopsies. }
    \label{fig:cm_all}
\end{figure}

\begin{figure}[H]
    \centering
    \includegraphics[width=\textwidth]{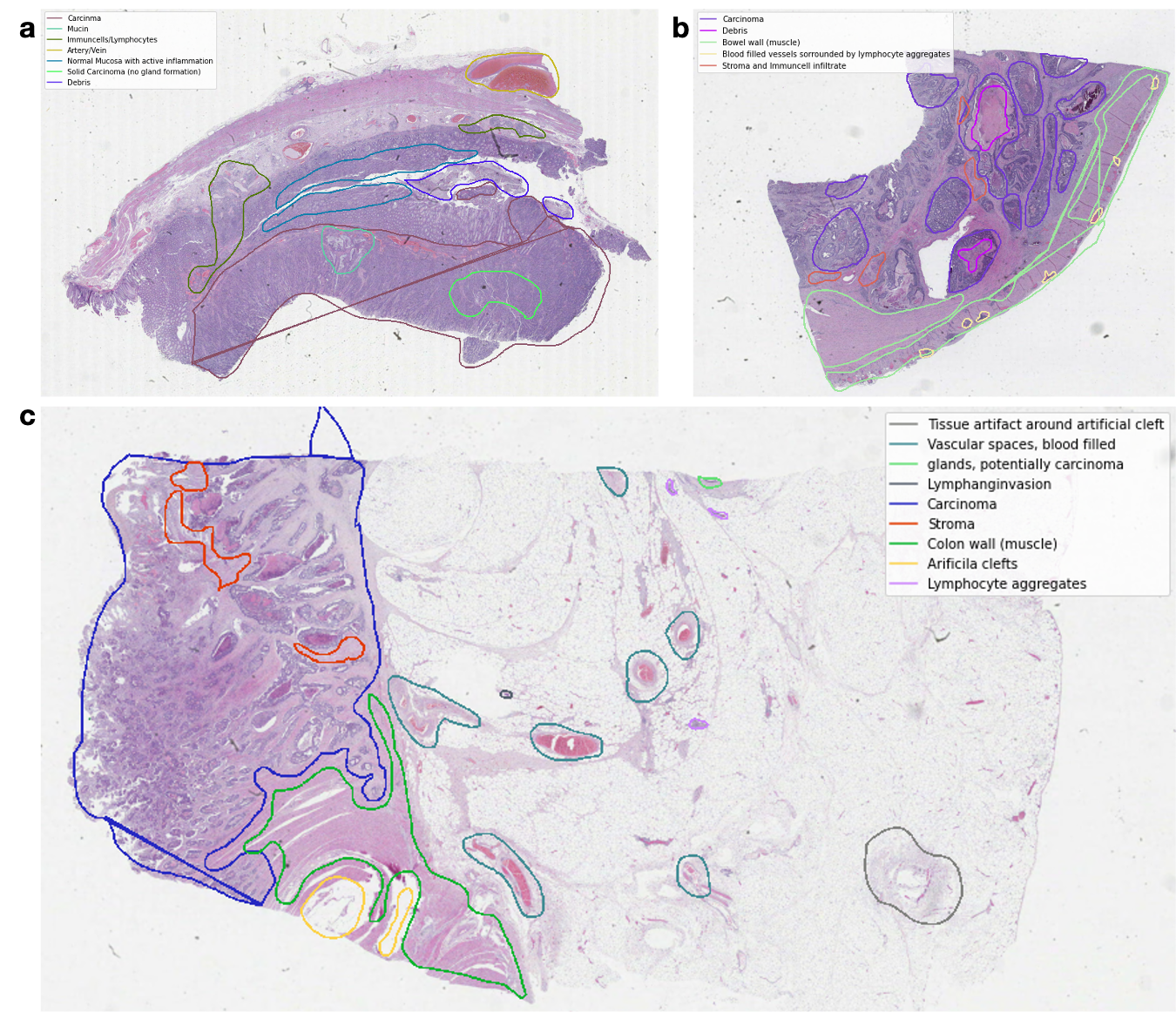}
    \caption{\textbf{Manual annotation by a pathologist of the three test cases used for attention visualization from the YCR-BCIP cohort.}}
    \label{fig:annotations}
\end{figure}


\subsection{Supplementary Tables}


\begin{sidewaystable}[ht]
    \protect\tiny
    \centering
    \caption{\textbf{Multi-cohort experiments with external validation on YCR-BCIP and YCR-BCIP-biopsies.} The models were trained with the same pre-processing and feature extractor, only varying the architecture of the aggregation model. All models were trained with 5-fold cross validation. }
    \label{tab:results}
	\begin{tabular}{lllllll|cccccccc}
        \hline    
        \textbf{ID} & \textbf{Train}            & \textbf{Test}             & \textbf{Target} & \makecell[l]{\textbf{Normali-}\\\textbf{zation}} & \makecell[l]{\textbf{Feature}\\\textbf{Extraction}} & \textbf{Aggregation Model}          & \makecell[c]{\textbf{AUROC}\\\textbf{mean}} & \makecell[c]{\textbf{AUROC}\\\textbf{std dev}} & \makecell[c]{\textbf{AUPRC}\\\textbf{mean}} & \makecell[c]{\textbf{AUPRC}\\\textbf{std dev}} & \makecell[c]{\textbf{F1 (0.5)}\\\textbf{mean}} & \makecell[c]{\textbf{F1 (0.5)}\\\textbf{std dev}} & \makecell[c]{\textbf{F1 (gmean)}\\\textbf{mean}}  & \makecell[c]{\textbf{F1 (gmean)}\\\textbf{std dev}} \\ \hline \hline
        {2.1.1} & \makecell[l]{DACHS, NLCS,\\QUASAR, TCGA } & \makecell[l]{DACHS, NLCS,\\QUASAR, TCGA } & {MSI-H} & {Macenko} & \makecell[l]{CTransPath\\\cite{Wang2022-fd}} & \makecell[l]{Transformer with class\\token (ours)} & {0.95} & {0.0078} & {0.74} & {0.0284} & {0.83} & {0.1257} & {0.80} & {0.1370} \\
        {2.2.1} & \makecell[l]{DACHS, NLCS,\\QUASAR, TCGA} & {YCR-BCIP} & {MSI-H} & {Macenko} & \makecell[l]{CTransPath\\\cite{Wang2022-fd}} & \makecell[l]{Transformer with class\\token (ours)} & \textbf{0.97} & {0.0041} & {0.83} & {0.0266} & \textbf{0.83} & {0.1145} & \textbf{0.84} & {0.1108} \\
        {2.3.1} & \makecell[l]{DACHS, NLCS,\\QUASAR, TCGA } & {YCR-BCIP-biopsies} & {MSI-H} & {Macenko} & \makecell[l]{CTransPath\\\cite{Wang2022-fd}} & \makecell[l]{Transformer with class\\token (ours)} & \textbf{0.91} & {0.0094} & {0.63} & {0.0149} & \textbf{0.77} & {0.1616} & \textbf{0.74} & {0.1622} \\
        \hline
        {2.1.2} & \makecell[l]{DACHS, NLCS,\\QUASAR, TCGA} & \makecell[l]{DACHS, NLCS,\\QUASAR, TCGA} & {MSI-H} & {Macenko} & \makecell[l]{CTransPath\\\cite{Wang2022-fd}} & \makecell[l]{Transformer with\\global averaging (ours)} & {0.95} & {0.0091} & {0.76} & {0.0224} & {0.83} & {0.1268} & {0.81} & {0.1322} \\ 
        {2.2.2} & \makecell[l]{DACHS, NLCS,\\QUASAR, TCGA} & {YCR-BCIP} & {MSI-H} & {Macenko} & \makecell[l]{CTransPath\\\cite{Wang2022-fd}} & \makecell[l]{Transformer with\\global averaging (ours)} & \textbf{0.97} & {0.0042} & \textbf{0.84} & {0.0131} & {0.82} & {0.1272} & {0.83} & {0.1191} \\
        {2.3.2} & \makecell[l]{DACHS, NLCS,\\QUASAR, TCGA} & {YCR-BCIP-biopsies} & {MSI-H} & {Macenko} & \makecell[l]{CTransPath\\\cite{Wang2022-fd}} & \makecell[l]{Transformer with\\global averaging (ours)} & \textbf{0.91} & {0.0078} & \textbf{0.64} & {0.0206} & {0.75} & {0.1795} & \textbf{0.74} & {0.1582} \\
        \hline
        {2.1.3} & \makecell[l]{DACHS, NLCS,\\QUASAR, TCGA} & \makecell[l]{DACHS, NLCS,\\QUASAR, TCGA} & {MSI-H} & {Macenko} & \makecell[l]{CTransPath\\\cite{Wang2022-fd}} & \makecell[l]{AttentionMIL\\\cite{Ilse2018-it}} & {0.94} & {0.0103} & {0.71} & {0.0184} & {0.79} & {0.1477} & {0.77} & {0.1543} \\
        {2.2.3} & \makecell[l]{DACHS, NLCS,\\QUASAR, TCGA} & {YCR-BCIP} & {MSI-H} & {Macenko} & \makecell[l]{CTransPath\\\cite{Wang2022-fd}} & \makecell[l]{AttentionMIL\\\cite{Ilse2018-it}} & {0.96} & {0.0025} & {0.80} & {0.0101} & {0.78} & {0.1413} & {0.82} & {0.1202} \\
        {2.3.3} & \makecell[l]{DACHS, NLCS,\\QUASAR, TCGA} & {YCR-BCIP-biopsies} & {MSI-H} & {Macenko} & \makecell[l]{CTransPath\\\cite{Wang2022-fd}} & \makecell[l]{AttentionMIL\\\cite{Ilse2018-it}} & {0.90} & {0.0042} & {0.60} & {0.0154} & {0.76} & {0.1601} & \textbf{0.74} & {0.1595} \\
        \hline
        {2.1.4} & \makecell[l]{DACHS, NLCS,\\QUASAR, TCGA} & \makecell[l]{DACHS, NLCS,\\QUASAR, TCGA} & {MSI-H} & {Macenko} & \makecell[l]{CTransPath\\\cite{Wang2022-fd}} & \makecell[l]{TransMILM\\\cite{Shao2021-vs}} & {0.94} & {0.0101} & {0.72} & {0.0379} & {0.82} & {0.1387} & {0.79} & {0.1500} \\
        {2.2.4} & \makecell[l]{DACHS, NLCS,\\QUASAR, TCGA} & {YCR-BCIP} & {MSI-H} & {Macenko} & \makecell[l]{CTransPath\\\cite{Wang2022-fd}} & \makecell[l]{TransMILM\\\cite{Shao2021-vs}} & {0.96} & {0.0033} & {0.79} & {0.0200} & {0.84} & {0.1157} & {0.83} & {0.1155} \\
        {2.3.4} & \makecell[l]{DACHS, NLCS,\\QUASAR, TCGA} & {YCR-BCIP-biopsies} & {MSI-H} & {Macenko} & \makecell[l]{CTransPath\\\cite{Wang2022-fd}} & \makecell[l]{TransMILM\\\cite{Shao2021-vs}} & {0.89} & {0.0122} & {0.57} & {0.0178} & {0.72} & {0.2215} & {0.70} & {0.1736} \\
        \hline
    \end{tabular}
\end{sidewaystable}

\begin{sidewaystable}
    \protect\scriptsize
    \caption{STARD (STAndards for the Reporting of Diagnostic accuracy studies) Checklist.}
    \label{tab:stard}
    \centering
    \begin{tabular}{l|l|l|c}
    \hline
    \textbf{Section \& Topic}  & \textbf{No}   & \textbf{Item}    & \textbf{Reported} \\ \hline \hline
    {TITLE OR ABSTRACT }& {1}   & {Identification as a study of diagnostic accuracy using at least one measure of accuracy (such as sensitivity, specificity, predictive values, or AUC)} & {yes} \\ \hline
    {ABSTRACT } & {2} & {Structured summary of study design, methods, results, and conclusions (for specific guidance, see STARD for Abstracts) } & {yes} \\ \hline
    {INTRODUCTION} & {3} & {Scientific and clinical background, including the intended use and clinical role of the index test} & {yes} \\ 
    {} & {4} & {Study objectives and hypotheses } & {yes} \\ \hline
    \makecell[l]{METHODS Study design} & {5} & {Whether data collection was planned before the index test and reference standard were performed (prospective study) or after (retrospective study)} & {} \\ \hline
    \makecell[l]{METHODS Participants} & {6} & {Eligibility criteria} & {} \\
    {} & {7} & {On what basis potentially eligible participants were identified (such as symptoms, results from previous tests, inclusion in registry) } & {} \\
    {} & {8} & {Where and when potentially eligible participants were identified (setting, location and dates) } & {} \\
    {} & {9} &  {Whether participants formed a consecutive, random or convenience series} & {} \\ \hline
    \makecell[l]{METHODS Test methods} & {10a} & {Index test, in sufficient detail to allow replication} & {yes} \\
    {} & {10b} & {Reference standard, in sufficient detail to allow replication } & {yes} \\
    {} & {11} & {Rationale for choosing the reference standard (if alternatives exist)} & {} \\
    {} & {12a} & {Definition of and rationale for test positivity cut-offs or result categories of the index test, distinguishing pre-specified from exploratory} & {yes} \\
    {} & {12b} & {Definition of and rationale for test positivity cut-offs or result categories of the reference standard, distinguishing pre-specified from exploratory } & {} \\
    {} & {13a} & {Whether clinical information and reference standard results were available to the performers/readers of the index test } & {} \\
    {} & {13b} & {Whether clinical information and index test results were available to the assessors of the reference standard } & {} \\ \hline
    \makecell[l]{METHODS Analysis} & {14} & {Methods for estimating or comparing measures of diagnostic accuracy} & {yes} \\
    {} & {15} & {How indeterminate index test or reference standard results were handled} & {} \\
    {} & {16} & {How missing data on the index test and reference standard were handled} & {yes} \\
    {} & {17} & {Any analyses of variability in diagnostic accuracy, distinguishing pre-specified from exploratory}  & \\
    {} & {18} & {Intended sample size and how it was determined } & {yes} \\
    \makecell[l]{RESULTS Participants} & {19} & {Flow of participants, using a diagram} & {} \\ \hline
    {} & {20} & {Baseline demographic and clinical characteristics of participants} & {yes} \\
    {} & {21a} & {Distribution of severity of disease in those with the target condition} & {} \\
    {} & {21b} & {Distribution of alternative diagnoses in those without the target condition} & {yes} \\ 
    {} & {22} & {Time interval and any clinical interventions between index test and reference standard } & {} \\ \hline
    \makecell[l]{RESULTS Test results } & {23} & {Cross tabulation of the index test results (or their distribution) by the results of the reference standard } & {} \\
    {} & {24} & {Estimates of diagnostic accuracy and their precision (such as 95\% confidence intervals) } & {yes} \\
    {} & {25} & {Any adverse events from performing the index test or the reference standard } & {} \\ \hline
    {DISCUSSION} & {26} & {Study limitations, including sources of potential bias, statistical uncertainty, and generalisability} & {yes} \\
    {} & {27} & {Implications for practice, including the intended use and clinical role of the index test } & {} \\ \hline
    {OTHER INFORMATION } & {28} & {Registration number and name of registry} & {} \\
    {} & {29} & {Where the full study protocol can be accessed} & {} \\
    {} & {30} & {Sources of funding and other support; role of funders} & {yes} \\ \hline
    \end{tabular}
\end{sidewaystable}

\begin{sidewaystable}
    \protect\tiny
    \caption{\textbf{Patient cohorts used in this study and their characteristics.} Clinico-pathological data were provided by the respective study principal investigators. In all cases, the TNM version from the original study registry was used. Information about the localization of the tumor was either provided as a binary variable (left-sided vs. right-sided) by the study site or assigned by the authors as follows: the cecum, ascending colon, hepatic flexure and transverse colon were defined as a right-sided tumor location whereas the splenic flexure, descending colon, sigmoid colon and rectum were defined as left-sided. *Number of patients before dropout of samples. **for the MECC cohort, these statistics refer to the cases with available MSI/dMMR status only.}
    \label{tab:cohorts}
    \centering
    \begin{tabular}{l|cccccccccc}
    \hline
\multicolumn{1}{c}{\textbf{}}         
    & \textbf{CPTAC}                                                   & \textbf{DACHS}                                                   & \textbf{DUSSEL}                                                  & \textbf{Epi700}                                                 & \textbf{MECC}**                                               & \textbf{MUNICH}                                                 & \textbf{NLCS}                                                   & \textbf{QUASAR}                                                 & \textbf{TCGA}                                                    & \textbf{YCR-BCIP}                                               \\ \hline \hline
Origin                                                                       & United States                                           & Germany                                                 & Germany                                                 & Northern Ireland                                       & Israel                                               & Germany                                                & Netherlands                                            & United Kingdom                                         & United States                                           & United Kingdom                                         \\
Number of patients*                                                          & 110                                                     & 2448                                                    & 330                                                     & 657                                                    & 683                                                  & 292                                                    & 2452                                                   & 2190                                                   & 632                                                     & 889                                                    \\
WSI format                                                                   & SVS                                                     & SVS                                                     & SVS                                                     & SVS                                                    & TIF                                                  & SVS                                                    & TIFF/SVS                                               & SVS                                                    & SVS                                                     & SVS                                                    \\ 
\hline
MSI-H/dMMR ground truth                                                      & MuTect2                                                 & PCR 3-plex                                              & IHC 2-plex                                              & PCR/IHC consensus                                      & PCR 5-plex                                           & IHC 4-plex                                             & IHC 2-plex                                             & IHC 4-plex or IHC 2-plex                               & PCR 5-plex                                              & IHC 4-plex                                             \\
\begin{tabular}[c]{@{}l@{}}MSI-H/dMMR, n \\ (\%)\end{tabular}                & \begin{tabular}[c]{@{}c@{}}24\\ (22\%)\end{tabular}     & \begin{tabular}[c]{@{}c@{}}210\\ (9\%)\end{tabular}     & \begin{tabular}[c]{@{}c@{}}45\\ (14\%)\end{tabular}     & \begin{tabular}[c]{@{}c@{}}134\\ (20\%)\end{tabular}   & \begin{tabular}[c]{@{}c@{}}106\\ (16\%)\end{tabular} & \begin{tabular}[c]{@{}c@{}}34\\ (12\%)\end{tabular}    & \begin{tabular}[c]{@{}c@{}}259\\ (11\%)\end{tabular}   & \begin{tabular}[c]{@{}c@{}}246\\ (11\%)\end{tabular}   & \begin{tabular}[c]{@{}c@{}}65\\ (10\%)\end{tabular}     & \begin{tabular}[c]{@{}c@{}}129\\ (15\%)\end{tabular}   \\
\begin{tabular}[c]{@{}l@{}}MSS/pMMR, n \\ (\%)\end{tabular}                  & \begin{tabular}[c]{@{}c@{}}81\\ (74\%)\end{tabular}     & \begin{tabular}[c]{@{}c@{}}1836\\ (75\%)\end{tabular}   & \begin{tabular}[c]{@{}c@{}}268\\ (81\%)\end{tabular}    & \begin{tabular}[c]{@{}c@{}}469\\ (71\%)\end{tabular}   & \begin{tabular}[c]{@{}c@{}}577\\ (84\%)\end{tabular} & \begin{tabular}[c]{@{}c@{}}258\\ (88\%)\end{tabular}   & \begin{tabular}[c]{@{}c@{}}2193\\ (89\%)\end{tabular}  & \begin{tabular}[c]{@{}c@{}}1529\\ (70\%)\end{tabular}  & \begin{tabular}[c]{@{}c@{}}392\\ (62\%)\end{tabular}    & \begin{tabular}[c]{@{}c@{}}760\\ (85\%)\end{tabular}   \\ 
\hline

\begin{tabular}[c]{@{}l@{}}Mean age at diagnosis \\ (std. dev.)\end{tabular} & \begin{tabular}[c]{@{}c@{}}65.67\\ (11.38)\end{tabular} & \begin{tabular}[c]{@{}c@{}}68.46\\ (10.82)\end{tabular} & \begin{tabular}[c]{@{}c@{}}68.57\\ (11.77)\end{tabular} & \begin{tabular}[c]{@{}c@{}}70.63\\ (11.4)\end{tabular} & 69.8                                                 & \begin{tabular}[c]{@{}c@{}}56.1\\ (11.84)\end{tabular} & \begin{tabular}[c]{@{}c@{}}73.71\\ (6.04)\end{tabular} & \begin{tabular}[c]{@{}c@{}}62.20\\ (9.60)\end{tabular} & \begin{tabular}[c]{@{}c@{}}66.42\\ (12.67)\end{tabular} & \begin{tabular}[c]{@{}c@{}}70.31\\ (9.97)\end{tabular} \\ 
\hline

\begin{tabular}[c]{@{}l@{}}Colon cancer, n \\ (\%)\end{tabular}              & \begin{tabular}[c]{@{}c@{}}110\\ (100\%)\end{tabular}   & \begin{tabular}[c]{@{}c@{}}1488\\ (61\%)\end{tabular}   & \begin{tabular}[c]{@{}c@{}}204\\  (62\%)\end{tabular}   & \begin{tabular}[c]{@{}c@{}}659\\ (99.7\%)\end{tabular} & \begin{tabular}[c]{@{}c@{}}530\\ (78\%)\end{tabular} & N/A                                                    & \begin{tabular}[c]{@{}c@{}}1730\\ (71\%)\end{tabular}  & \begin{tabular}[c]{@{}c@{}}1474\\ (67\%)\end{tabular}  & \begin{tabular}[c]{@{}c@{}}341\\ (54\%)\end{tabular}    & \begin{tabular}[c]{@{}c@{}}667\\ (75\%)\end{tabular}   \\
\begin{tabular}[c]{@{}l@{}}Rectal cancer, n \\ (\%)\end{tabular}             & \begin{tabular}[c]{@{}c@{}}0\\ (0\%)\end{tabular}       & \begin{tabular}[c]{@{}c@{}}960\\ (39\%)\end{tabular}    & \begin{tabular}[c]{@{}c@{}}116\\ (35\%)\end{tabular}    & \begin{tabular}[c]{@{}c@{}}2\\ (0.3\%)\end{tabular}    & \begin{tabular}[c]{@{}c@{}}123\\ (18\%)\end{tabular} & N/A                                                    & \begin{tabular}[c]{@{}c@{}}722\\ (29\%)\end{tabular}   & \begin{tabular}[c]{@{}c@{}}526\\ (24\%)\end{tabular}   & \begin{tabular}[c]{@{}c@{}}118\\ (19\%)\end{tabular}    & \begin{tabular}[c]{@{}c@{}}215\\ (24\%)\end{tabular}   \\
\begin{tabular}[c]{@{}l@{}}Site unknown, n \\ (\%)\end{tabular}              & \begin{tabular}[c]{@{}c@{}}0\\ (0\%)\end{tabular}       & \begin{tabular}[c]{@{}c@{}}0 \\ (0\%)\end{tabular}      & \begin{tabular}[c]{@{}c@{}}10\\ (3\%)\end{tabular}      & \begin{tabular}[c]{@{}c@{}}0\\ (0\%)\end{tabular}      & \begin{tabular}[c]{@{}c@{}}30\\ (4\%)\end{tabular}   & N/A                                                    & \begin{tabular}[c]{@{}c@{}}0\\ (0\%)\end{tabular}      & \begin{tabular}[c]{@{}c@{}}190\\ (9\%)\end{tabular}    & \begin{tabular}[c]{@{}c@{}}173\\  (27\%)\end{tabular}   & \begin{tabular}[c]{@{}c@{}}7\\ (1\%)\end{tabular}      \\
\hline

\begin{tabular}[c]{@{}l@{}}Female, n \\ (\%)\end{tabular}                    & \begin{tabular}[c]{@{}c@{}}65\\ (59\%)\end{tabular}     & \begin{tabular}[c]{@{}c@{}}1012\\ (41\%)\end{tabular}   & \begin{tabular}[c]{@{}c@{}}181\\ (55\%)\end{tabular}    & \begin{tabular}[c]{@{}c@{}}303\\ (46\%)\end{tabular}   & \begin{tabular}[c]{@{}c@{}}320\\ (47\%)\end{tabular} & \begin{tabular}[c]{@{}c@{}}132\\ (45\%)\end{tabular}   & \begin{tabular}[c]{@{}c@{}}1079\\ (44\%)\end{tabular}  & \begin{tabular}[c]{@{}c@{}}848\\ (39\%)\end{tabular}   & \begin{tabular}[c]{@{}c@{}}292\\ (46\%)\end{tabular}    & \begin{tabular}[c]{@{}c@{}}395\\ (44\%)\end{tabular}   \\
\begin{tabular}[c]{@{}l@{}}Male, n \\ (\%)\end{tabular}                      & \begin{tabular}[c]{@{}c@{}}45\\ (41\%)\end{tabular}     & \begin{tabular}[c]{@{}c@{}}1436\\ (59\%)\end{tabular}   & \begin{tabular}[c]{@{}c@{}}149\\ (45\%)\end{tabular}    & \begin{tabular}[c]{@{}c@{}}358\\ (54\%)\end{tabular}   & \begin{tabular}[c]{@{}c@{}}363\\ (53\%)\end{tabular} & \begin{tabular}[c]{@{}c@{}}160\\ (55\%)\end{tabular}   & \begin{tabular}[c]{@{}c@{}}1373\\ (56\%)\end{tabular}  & \begin{tabular}[c]{@{}c@{}}1334\\ (61\%)\end{tabular}  & \begin{tabular}[c]{@{}c@{}}322\\ (51\%)\end{tabular}    & \begin{tabular}[c]{@{}c@{}}494\\ (56\%)\end{tabular}   \\
\begin{tabular}[c]{@{}l@{}}gender unknown, n\\ (\%)\end{tabular}             & \begin{tabular}[c]{@{}c@{}}0\\ (0\%)\end{tabular}       & \begin{tabular}[c]{@{}c@{}}0 \\ (0\%)\end{tabular}      & \begin{tabular}[c]{@{}c@{}}0\\ (0\%)\end{tabular}       & \begin{tabular}[c]{@{}c@{}}0\\ (0\%)\end{tabular}      & \begin{tabular}[c]{@{}c@{}}0\\ (0\%)\end{tabular}    & \begin{tabular}[c]{@{}c@{}}0 \\ (0\%)\end{tabular}     & \begin{tabular}[c]{@{}c@{}}0\\ (0\%)\end{tabular}      & \begin{tabular}[c]{@{}c@{}}8\\ (0\%)\end{tabular}      & \begin{tabular}[c]{@{}c@{}}18\\ (3\%)\end{tabular}      & \begin{tabular}[c]{@{}c@{}}0\\ (0\%)\end{tabular}      \\
\hline

\begin{tabular}[c]{@{}l@{}}UICC stage I, n \\ (\%)\end{tabular}              & \begin{tabular}[c]{@{}c@{}}12\\ (11\%)\end{tabular}     & \begin{tabular}[c]{@{}c@{}}485\\ (20\%)\end{tabular}    & \begin{tabular}[c]{@{}c@{}}76\\ (23\%)\end{tabular}     & \begin{tabular}[c]{@{}c@{}}0\\ (0\%)\end{tabular}      & \begin{tabular}[c]{@{}c@{}}94 \\ (14\%)\end{tabular} & \begin{tabular}[c]{@{}c@{}}52\\ (18\%)\end{tabular}    & \begin{tabular}[c]{@{}c@{}}485\\ (20\%)\end{tabular}   & \begin{tabular}[c]{@{}c@{}}1\\  (0\%)\end{tabular}     & \begin{tabular}[c]{@{}c@{}}76\\ (12\%)\end{tabular}     & \begin{tabular}[c]{@{}c@{}}169\\ (19\%)\end{tabular}   \\
\begin{tabular}[c]{@{}l@{}}UICC stage II, n \\ (\%)\end{tabular}             & \begin{tabular}[c]{@{}c@{}}42\\ (38\%)\end{tabular}     & \begin{tabular}[c]{@{}c@{}}801\\ (33\%)\end{tabular}    & \begin{tabular}[c]{@{}c@{}}138\\ (42\%)\end{tabular}    & \begin{tabular}[c]{@{}c@{}}394\\ (60\%)\end{tabular}   & \begin{tabular}[c]{@{}c@{}}335\\ (49\%)\end{tabular} & \begin{tabular}[c]{@{}c@{}}118\\ (40\%)\end{tabular}   & \begin{tabular}[c]{@{}c@{}}918\\ (37\%)\end{tabular}   & \begin{tabular}[c]{@{}c@{}}1988\\ (91\%)\end{tabular}  & \begin{tabular}[c]{@{}c@{}}166\\ (26\%)\end{tabular}    & \begin{tabular}[c]{@{}c@{}}317\\ (36\%)\end{tabular}   \\
\begin{tabular}[c]{@{}l@{}}UICC stage III, n \\ (\%)\end{tabular}            & \begin{tabular}[c]{@{}c@{}}48\\ (44\%)\end{tabular}     & \begin{tabular}[c]{@{}c@{}}822\\ (34\%)\end{tabular}    & \begin{tabular}[c]{@{}c@{}}110\\ (33\%)\end{tabular}    & \begin{tabular}[c]{@{}c@{}}267\\ (40\%)\end{tabular}   & \begin{tabular}[c]{@{}c@{}}123\\ (18\%)\end{tabular} & \begin{tabular}[c]{@{}c@{}}82\\ (28\%)\end{tabular}    & \begin{tabular}[c]{@{}c@{}}641\\ (26\%)\end{tabular}   & \begin{tabular}[c]{@{}c@{}}192\\ (9\%)\end{tabular}    & \begin{tabular}[c]{@{}c@{}}140\\ (22\%)\end{tabular}    & \begin{tabular}[c]{@{}c@{}}370\\ (42\%)\end{tabular}   \\
\begin{tabular}[c]{@{}l@{}}UICC stage IV, n \\ (\%)\end{tabular}             & \begin{tabular}[c]{@{}c@{}}8\\ (7\%)\end{tabular}       & \begin{tabular}[c]{@{}c@{}}337\\ (14\%)\end{tabular}    & \begin{tabular}[c]{@{}c@{}}6\\ (2\%)\end{tabular}       & \begin{tabular}[c]{@{}c@{}}0\\ (0\%)\end{tabular}      & \begin{tabular}[c]{@{}c@{}}67\\ (10\%)\end{tabular}  & \begin{tabular}[c]{@{}c@{}}39\\ (13\%)\end{tabular}    & \begin{tabular}[c]{@{}c@{}}341\\ (14\%)\end{tabular}   & \begin{tabular}[c]{@{}c@{}}0\\ (0\%)\end{tabular}      & \begin{tabular}[c]{@{}c@{}}63\\ (10\%)\end{tabular}     & \begin{tabular}[c]{@{}c@{}}0\\ (0\%)\end{tabular}      \\
\begin{tabular}[c]{@{}l@{}}UICC stage unknown, n \\ (\%)\end{tabular}        & \begin{tabular}[c]{@{}c@{}}0\\ (0\%)\end{tabular}       & \begin{tabular}[c]{@{}c@{}}3\\ (0\%)\end{tabular}       & \begin{tabular}[c]{@{}c@{}}0\\ (0\%)\end{tabular}       & \begin{tabular}[c]{@{}c@{}}0\\ (0\%)\end{tabular}      & \begin{tabular}[c]{@{}c@{}}64\\ (9\%)\end{tabular}   & \begin{tabular}[c]{@{}c@{}}1\\ (0\%)\end{tabular}      & \begin{tabular}[c]{@{}c@{}}67\\ (3\%)\end{tabular}     & \begin{tabular}[c]{@{}c@{}}9\\ (0\%)\end{tabular}      & \begin{tabular}[c]{@{}c@{}}187\\ (30\%)\end{tabular}    & \begin{tabular}[c]{@{}c@{}}33\\ (3\%)\end{tabular}     \\
\hline

\begin{tabular}[c]{@{}l@{}}\textit{BRAF} mutation, n \\ (\%)\end{tabular}             & N/A                                                     & \begin{tabular}[c]{@{}c@{}}151\\ (6\%)\end{tabular}     & N/A                                                     & \begin{tabular}[c]{@{}c@{}}91\\ (14\%)\end{tabular}    & \begin{tabular}[c]{@{}c@{}}49\\ (7\%)\end{tabular}   & N/A                                                    & \begin{tabular}[c]{@{}c@{}}305\\ (12\%)\end{tabular}   & \begin{tabular}[c]{@{}c@{}}120\\ (5\%)\end{tabular}    & \begin{tabular}[c]{@{}c@{}}63\\ (10\%)\end{tabular}     & \begin{tabular}[c]{@{}c@{}}75\\ (8\%)\end{tabular}     \\
\begin{tabular}[c]{@{}l@{}}\textit{BRAF} wild type, n\\ (\%)\end{tabular}             & N/A                                                     & \begin{tabular}[c]{@{}c@{}}1930\\ (79\%)\end{tabular}   & N/A                                                     & \begin{tabular}[c]{@{}c@{}}550\\ (84\%)\end{tabular}   & \begin{tabular}[c]{@{}c@{}}570\\ (83\%)\end{tabular} & N/A                                                    & \begin{tabular}[c]{@{}c@{}}1733\\ (71\%)\end{tabular}  & \begin{tabular}[c]{@{}c@{}}1358\\ (62\%)\end{tabular}  & \begin{tabular}[c]{@{}c@{}}471\\ (75\%)\end{tabular}    & \begin{tabular}[c]{@{}c@{}}32\\ (4\%)\end{tabular}     \\
\begin{tabular}[c]{@{}l@{}}\textit{BRAF} status unknown,  n \\ (\%)\end{tabular}      & N/A                                                     & \begin{tabular}[c]{@{}c@{}}367\\ (15\%)\end{tabular}    & N/A                                                     & \begin{tabular}[c]{@{}c@{}}16\\ (2\%)\end{tabular}     & \begin{tabular}[c]{@{}c@{}}64\\ (9\%)\end{tabular}   & N/A                                                    & \begin{tabular}[c]{@{}c@{}}414\\ (17\%)\end{tabular}   & \begin{tabular}[c]{@{}c@{}}712\\ (33\%)\end{tabular}   & \begin{tabular}[c]{@{}c@{}}98\\  (15\%)\end{tabular}    & \begin{tabular}[c]{@{}c@{}}782\\ (88\%)\end{tabular}   \\
\hline

\begin{tabular}[c]{@{}l@{}}\textit{KRAS} mutation, n \\ (\%)\end{tabular}             & N/A                                                     & \begin{tabular}[c]{@{}c@{}}677\\ (28\%)\end{tabular}    & N/A                                                     & \begin{tabular}[c]{@{}c@{}}247\\ (38\%)\end{tabular}   & \begin{tabular}[c]{@{}c@{}}252\\ (37\%)\end{tabular} & N/A                                                    & \begin{tabular}[c]{@{}c@{}}698\\ (28\%)\end{tabular}   & \begin{tabular}[c]{@{}c@{}}555\\ (25\%)\end{tabular}   & \begin{tabular}[c]{@{}c@{}}218\\ (34\%)\end{tabular}    & N/A                                                    \\
\begin{tabular}[c]{@{}l@{}}\textit{KRAS} wild type, n \\ (\%)\end{tabular}            & N/A                                                     & \begin{tabular}[c]{@{}c@{}}1397\\ (57\%)\end{tabular}   & N/A                                                     & \begin{tabular}[c]{@{}c@{}}398\\ (61\%)\end{tabular}   & \begin{tabular}[c]{@{}c@{}}405\\ (59\%)\end{tabular} & N/A                                                    & \begin{tabular}[c]{@{}c@{}}1335\\ (54\%)\end{tabular}  & \begin{tabular}[c]{@{}c@{}}882\\ (40\%)\end{tabular}   & \begin{tabular}[c]{@{}c@{}}316\\ (50\%)\end{tabular}    & N/A                                                    \\
\begin{tabular}[c]{@{}l@{}}\textit{KRAS} status unknown, n\\ (\%)\end{tabular}        & N/A                                                     & \begin{tabular}[c]{@{}c@{}}347\\ (15\%)\end{tabular}    & N/A                                                     & \begin{tabular}[c]{@{}c@{}}12\\ (2\%)\end{tabular}     & \begin{tabular}[c]{@{}c@{}}26\\ (4\%)\end{tabular}   & N/A                                                    & \begin{tabular}[c]{@{}c@{}}419\\ (17\%)\end{tabular}   & \begin{tabular}[c]{@{}c@{}}753\\ (35\%)\end{tabular}   & \begin{tabular}[c]{@{}c@{}}98\\ (16\%)\end{tabular}     & N/A                                                    \\
\hline

\begin{tabular}[c]{@{}l@{}}right-sided tumor, n \\ (\%)\end{tabular}         & \begin{tabular}[c]{@{}c@{}}58\\ (53\%)\end{tabular}     & \begin{tabular}[c]{@{}c@{}}819\\ (33\%)\end{tabular}    & \begin{tabular}[c]{@{}c@{}}72\\ (22\%)\end{tabular}     & \begin{tabular}[c]{@{}c@{}}375\\ (57\%)\end{tabular}   & \begin{tabular}[c]{@{}c@{}}238\\ (35\%)\end{tabular} & \begin{tabular}[c]{@{}c@{}}53\\ (18\%)\end{tabular}    & \begin{tabular}[c]{@{}c@{}}946\\ (39\%)\end{tabular}   & \begin{tabular}[c]{@{}c@{}}754\\ (34\%)\end{tabular}   & \begin{tabular}[c]{@{}c@{}}176\\ (28\%)\end{tabular}    & \begin{tabular}[c]{@{}c@{}}331\\ (37\%)\end{tabular}   \\
\begin{tabular}[c]{@{}l@{}}left-sided tumor, n \\ (\%)\end{tabular}          & \begin{tabular}[c]{@{}c@{}}51\\ (46\%)\end{tabular}     & \begin{tabular}[c]{@{}c@{}}1607\\ (66\%)\end{tabular}   & \begin{tabular}[c]{@{}c@{}}226\\ (68\%)\end{tabular}    & \begin{tabular}[c]{@{}c@{}}280\\ (42\%)\end{tabular}   & \begin{tabular}[c]{@{}c@{}}409\\ (60\%)\end{tabular} & \begin{tabular}[c]{@{}c@{}}239\\ (82\%)\end{tabular}   & \begin{tabular}[c]{@{}c@{}}1506\\ (61\%)\end{tabular}  & \begin{tabular}[c]{@{}c@{}}1158\\ (53\%)\end{tabular}  & \begin{tabular}[c]{@{}c@{}}248\\ (39\%)\end{tabular}    & \begin{tabular}[c]{@{}c@{}}486\\ (55\%)\end{tabular}   \\
\begin{tabular}[c]{@{}l@{}}sidedness unknown, n \\ (\%)\end{tabular}         & \begin{tabular}[c]{@{}c@{}}1\\ (1\%)\end{tabular}       & \begin{tabular}[c]{@{}c@{}}22\\ (1\%)\end{tabular}      & \begin{tabular}[c]{@{}c@{}}32\\ (10\%)\end{tabular}     & \begin{tabular}[c]{@{}c@{}}6\\ (1\%)\end{tabular}      & \begin{tabular}[c]{@{}c@{}}36\\ (5\%)\end{tabular}   & \begin{tabular}[c]{@{}c@{}}0 \\ (0\%)\end{tabular}     & \begin{tabular}[c]{@{}c@{}}0 \\ (0\%)\end{tabular}     & \begin{tabular}[c]{@{}c@{}}150\\ (13\%)\end{tabular}   & \begin{tabular}[c]{@{}c@{}}208\\ (33\%)\end{tabular}    & \begin{tabular}[c]{@{}c@{}}72\\ (8\%)\end{tabular}     \\
\hline

etiology                                                                     & not specified                                           & any                                                     & not specified                                           & not specified                                          & not specified                                        & not specified                                          & not specified                                          & not specified                                          & not specified                                           & not specified \\ \hline
\end{tabular}
\end{sidewaystable}

\begin{sidewaystable}[h]
    \protect\tiny
    \centering
    \caption{Multi-cohort experiments for BRAF and KRAS prediction with external validation on Epi700. All models were trained with 5-fold cross validation. }
    \label{tab:braf_kras}
    \begin{tabular}{lllllll|cccccccc}
    \hline
    \textbf{ID} & \textbf{Train}            & \textbf{Test}             & \textbf{Target} & \makecell[l]{\textbf{Normali-}\\\textbf{zation}} & \makecell[l]{\textbf{Feature}\\\textbf{Extraction}} & \textbf{Aggregation Model}          & \makecell[c]{\textbf{AUROC}\\\textbf{mean}} & \makecell[c]{\textbf{AUROC}\\\textbf{std dev}} & \makecell[c]{\textbf{AUPRC}\\\textbf{mean}} & \makecell[c]{\textbf{AUPRC}\\\textbf{std dev}} & \makecell[c]{\textbf{F1 (0.5)}\\\textbf{mean}} & \makecell[c]{\textbf{F1 (0.5)}\\\textbf{std dev}} & \makecell[c]{\textbf{F1 (gmean)}\\\textbf{mean}}  & \makecell[c]{\textbf{F1 (gmean)}\\\textbf{std dev}} \\
    
    \hline \hline
    3.1.1           & \makecell[l]{DACHS, NLCS,\\QUASAR, TCGA } & \makecell[l]{DACHS, NLCS,\\QUASAR, TCGA } & \textit{BRAF}   & Macenko                 & \makecell[l]{CTransPath\\\cite{Wang2022-fd}}
                & \makecell[l]{Transformer with class\\token (ours)} & 0.86                                    & 0.0083                                     & 0.48                                    & 0.0241                                     & 0.69                                       & 0.2183                                        & 0.67                                         & 0.2093                                          \\
    3.2.1           & \makecell[l]{DACHS, NLCS,\\QUASAR, TCGA } & Epi700                    & \textit{BRAF}   & Macenko                 & \makecell[l]{CTransPath\\\cite{Wang2022-fd}}                & \makecell[l]{Transformer with class\\token (ours)} & 0.86                                    & 0.0130                                     & 0.53                                    & 0.0387                                     & 0.73                                       & 0.1754                                        & 0.72                                         & 0.1678                                          \\
    3.3.1           & \makecell[l]{DACHS, NLCS,\\QUASAR, TCGA } & \makecell[l]{DACHS, NLCS,\\QUASAR, TCGA } & \textit{KRAS}   & Macenko                 & \makecell[l]{CTransPath\\\cite{Wang2022-fd}}                & \makecell[l]{Transformer with class\\token (ours)} & 0.69                                    & 0.0162                                     & 0.54                                    & 0.0119                                     & 0.64                                       & 0.1332                                        & 0.64                                         & 0.0832                                          \\
    3.1.2           & \makecell[l]{DACHS, NLCS,\\QUASAR, TCGA } & Epi700                    & \textit{KRAS}   & Macenko                 & \makecell[l]{CTransPath\\\cite{Wang2022-fd}}                & \makecell[l]{Transformer with class\\token (ours)} & 0.75                                    & 0.0155                                     & 0.65                                    & 0.0209                                     & 0.67                                       & 0.0592                                        & 0.70                                         & 0.0577                       \\
    \hline
    \end{tabular}

    \vspace{12pt}
    \caption{Mean AUROC scores of 5-fold CV training on single cohorts, evaluated on all other cohorts. The training cohorts are listed in the columns and entries in the diagonal are in-domain test results. }
    \label{tab:single_cohort}
    \begin{tabular}{l|cccccccccc}
    \hline
    \textbf{Training cohort ($\downarrow$)} & \multicolumn{1}{l}{\textbf{NLCS}} & \multicolumn{1}{l}{\textbf{Dachs}} & \multicolumn{1}{l}{\textbf{Quasar}} & \multicolumn{1}{l}{\textbf{YCR-BCIP}} & \multicolumn{1}{l}{\textbf{MECC}} & \multicolumn{1}{l}{\textbf{Epi700}} & \multicolumn{1}{l}{\textbf{TCGA}} & \multicolumn{1}{l}{\textbf{Munich}} & \multicolumn{1}{l}{\textbf{Duessel}} & \multicolumn{1}{l}{\textbf{CPTAC}} \\ \hline \hline
    \textbf{NLCS}                & 0.93	& 0.92	& 0.93	& 0.96	& 0.79	& 0.93	& 0.88	& 0.86	& 0.80	& 0.90 \\
    \textbf{Dachs}               & 0.92	& 0.96	& 0.92	& 0.93	& 0.76	& 0.91	& 0.87	& 0.84	& 0.79	& 0.88 \\
    \textbf{Quasar}              & 0.92	& 0.92	& 0.95	& 0.95	& 0.76	& 0.93	& 0.88	& 0.88	& 0.81	& 0.91 \\
    \textbf{YCR-BCIP}            & 0.91	& 0.89	& 0.92	& 0.96	& 0.76	& 0.91	& 0.88	& 0.87	& 0.83	& 0.91 \\ 
    \textbf{MECC}                & 0.84	& 0.82	& 0.81	& 0.88	& 0.73	& 0.82	& 0.78	& 0.73	& 0.79	& 0.80 \\
    \textbf{Epi700}              & 0.90	& 0.87	& 0.88	& 0.92	& 0.76	& 0.90	& 0.85	& 0.84	& 0.76	& 0.85 \\
    \textbf{TCGA}                & 0.83	& 0.84	& 0.85	& 0.89	& 0.75	& 0.84	& 0.86	& 0.82	& 0.75	& 0.87 \\
    \textbf{Munich}              & 0.86	& 0.86	& 0.87	& 0.87	& 0.74	& 0.85	& 0.82	& 0.85	& 0.78	& 0.75 \\
    \textbf{Duessel}             & 0.76	& 0.79	& 0.79	& 0.85	& 0.67	& 0.75	& 0.73	& 0.74	& 0.78	& 0.73 \\
    \textbf{CPTAC}               & 0.78	& 0.76	& 0.75	& 0.83	& 0.68	& 0.88	& 0.74	& 0.70	& 0.73	& 0.81 \\ 
    \end{tabular}    

    \vspace{12pt}

    \protect\tiny
    \caption{Data for the performance analysis with a varying number of patients in the training set. The experiments were repeated five times and we report mean and standard deviation.}
    \label{tab:num_samples}
    \centering
\begin{tabular}{lccc|cc|cc|cc}
\hline
\textbf{Model →}             & \multicolumn{1}{l}{\textbf{}} & \multicolumn{2}{c}{\textbf{Transformer (ours)}}                                                                                          & \multicolumn{2}{c}{\textbf{AttentionMIL}}                                                                                                & \multicolumn{2}{c}{\textbf{Transformer (ours)}}                                                                                          & \multicolumn{2}{c}{\textbf{AttentionMIL}}                                                                                                \\
\textbf{Evaluation cohort →} & \multicolumn{1}{l}{}          & \multicolumn{2}{c}{\textbf{YCR-BCIP}}                                                                                                    & \multicolumn{2}{c}{\textbf{YCR-BCIP}}                                                                                                    & \multicolumn{2}{c}{\textbf{YCR-BCIP-biopsies}}                                                                                           & \multicolumn{2}{c}{\textbf{YCR-BCIP-biopsies}}                                                                                           \\ \hline
                             & \textbf{\#patients}           & \textbf{\begin{tabular}[c]{@{}c@{}}AUROC \\ (mean)\end{tabular}} & \textbf{\begin{tabular}[c]{@{}c@{}}AUROC \\ (std. dev.)\end{tabular}} & \textbf{\begin{tabular}[c]{@{}c@{}}AUROC \\ (mean)\end{tabular}} & \textbf{\begin{tabular}[c]{@{}c@{}}AUROC \\ (std. dev.)\end{tabular}} & \textbf{\begin{tabular}[c]{@{}c@{}}AUROC \\ (mean)\end{tabular}} & \textbf{\begin{tabular}[c]{@{}c@{}}AUROC \\ (std. dev.)\end{tabular}} & \textbf{\begin{tabular}[c]{@{}c@{}}AUROC \\ (mean)\end{tabular}} & \textbf{\begin{tabular}[c]{@{}c@{}}AUROC \\ (std. dev.)\end{tabular}} \\
    \hline \hline
                             & 50                            & 0.792                                                            & 0.0866                                                                & 0.616                                                            & 0.1252                                                                & 0.629                                                            & 0.0300                                                                & 0.554                                                            & 0.0640                                                                \\
                             & 100                           & 0.840                                                            & 0.0180                                                                & 0.744                                                            & 0.0546                                                                & 0.694                                                            & 0.0628                                                                & 0.604                                                            & 0.0518                                                                \\
                             & 250                           & 0.923                                                            & 0.0143                                                                & 0.822                                                            & 0.0281                                                                & 0.811                                                            & 0.0325                                                                & 0.653                                                            & 0.0372                                                                \\
                             & 500                           & 0.924                                                            & 0.0173                                                                & 0.838                                                            & 0.0344                                                                & 0.823                                                            & 0.0306                                                                & 0.673                                                            & 0.0356                                                                \\
                             & 1000                          & 0.944                                                            & 0.0148                                                                & 0.854                                                            & 0.0257                                                                & 0.853                                                            & 0.0441                                                                & 0.695                                                            & 0.0235                                                                \\
                             & 1500                          & 0.951                                                            & 0.0130                                                                & 0.865                                                            & 0.0134                                                                & 0.869                                                            & 0.0175                                                                & 0.716                                                            & 0.0273                                                                \\
                             & 2000                          & 0.946                                                            & 0.0046                                                                & 0.881                                                            & 0.0117                                                                & 0.882                                                            & 0.0213                                                                & 0.737                                                            & 0.0201                                                                \\
                             & 3000                          & 0.954                                                            & 0.0066                                                                & 0.890                                                            & 0.0057                                                                & 0.888                                                            & 0.0122                                                                & 0.755                                                            & 0.0217                                                                \\
                             & 4000                          & 0.950                                                            & 0.0065                                                                & 0.901                                                            & 0.0050                                                                & 0.877                                                            & 0.0151                                                                & 0.775                                                            & 0.0122                                                                \\
                             & 5000                          & 0.955                                                            & 0.0056                                                                & 0.906                                                            & 0.0027                                                                & 0.875                                                            & 0.0119                                                                & 0.789                                                            & 0.0101                                                                \\
                             & 6000                          & 0.962                                                            & 0.0096                                                                & 0.913                                                            & 0.0049                                                                & 0.900                                                            & 0.0155                                                                & 0.810                                                            & 0.0083                                                                \\
                             & 7000                          & 0.961                                                            & 0.0027                                                                & 0.918                                                            & 0.0028                                                                & 0.899                                                            & 0.0034                                                                & 0.823                                                            & 0.0044                                                                \\
                             & 8000                          & 0.960                                                            & 0.0080                                                                & 0.922                                                            & 0.0029                                                                & 0.902                                                            & 0.0221                                                                & 0.836                                                            & 0.0045                                            \\ \hline                   
\end{tabular}
\end{sidewaystable}

\end{document}